%% file: main.tex
\documentclass[10pt,twocolumn,letterpaper]{article}

\usepackage[pagenumbers]{cvpr} 

\input{preamble}

\definecolor{cvprblue}{rgb}{0.21,0.49,0.74}
\usepackage[pagebackref,breaklinks,colorlinks,allcolors=cvprblue]{hyperref}

\usepackage{utfsym}
\usepackage{multirow}
\usepackage[accsupp]{axessibility}  

\def\paperID{5046} 
\def\confName{CVPR}
\def\confYear{2025}

\title{ARKit LabelMaker: A New Scale for Indoor 3D Scene Understanding}

\author{}

\begin{document}

\twocolumn[{
            \renewcommand\twocolumn[1][]{#1}%
            \maketitle
            \vspace{-55px}
            \begin{center}
                {\large
                    Guangda Ji$^1$ \hspace{15px}
                    Silvan Weder$^1$ \hspace{15px}
                    Francis Engelmann$^{1,2}$ \hspace{15px}
                    Marc Pollefeys$^1$ \hspace{15px}
                    Hermann Blum$^{1,3}$ \hspace{15px}
                }\\
                \vspace{10px}
                $^1$ETH Zürich \hspace{15px}
                $^2$Stanford University \hspace{15px}
                $^3$University of Bonn \& Lamarr Institute \hspace{15px}
                \\
                \vspace{15px}
            \end{center}
        }]

\input{sec/0_abstract}
\input{sec/1_intro}

\input{sec/2_related_works}

\input{sec/3_method}

\input{sec/4_results}
\input{sec/5_conclusion}

\paragraph{Acknowledgements.}
We thank Xiaoyang Wu, author of PointTransformerV3~\cite{wu2024ptv3}, for valuable and helpful advice.
This project is partially supported by an SNSF Postdoc.Mobility fellowship, the ETH Foundation through a CareerSeed grant, and the Lamarr Institute.

\clearpage
{
    \small
    \bibliographystyle{ieeenat_fullname}
    \bibliography{main}
}

\makeatletter
\@ifpackagewith{cvpr}{pagenumbers}{
    \input{sec/X_suppl}

}{}
\makeatother

\end{document}

%% file: preamble.tex
\usepackage[skip=4px]{caption}
\renewcommand{\paragraph}[1]{\vspace{4px}\noindent\textbf{#1}\,}

%% file: sec/0_abstract.tex
\begin{abstract}
Neural network performance scales with both model size and data volume, as shown in both language and image processing. 
This requires scaling-friendly architectures and large datasets.
While transformers have been adapted for 3D vision,
a `GPT-moment' remains elusive due to limited training data.
We introduce ARKit LabelMaker,
a large-scale real-world 3D dataset with dense semantic annotation that is more than three times larger than prior largest dataset.
Specifically, we extend ARKitScenes~\cite{baruch2021arkitscenes} with automatically generated dense 3D labels using an extended LabelMaker pipeline~\cite{labelmaker}, tailored for large-scale pre-training. Training on our dataset improves accuracy across architectures, achieving state-of-the-art 3D semantic segmentation scores on ScanNet and ScanNet200, with notable gains on tail classes.
Our code is available at \href{https://labelmaker.org/}{labelmaker.org} and our dataset at \href{https://huggingface.co/datasets/labelmaker/arkit_labelmaker}{huggingface}.
\end{abstract}

%% file: sec/1_intro.tex
\section{Introduction}
\label{sec:intro}

Recent advancements in deep learning on language~\cite{radford2018gpt1,radford2019gpt2,brown2020gpt3} and 2D vision ~\cite{radford2021learning,rombach2022stablediffusion,achiam2023gpt} have made tremendous progress, primarily driven by the abundance of training data available on the web for these modalities.
Scaling this type of large-scale training to billions of data points has revealed surprising properties~\cite{wei2022emergent} and resulted in unprecedented performance gains, enabling entirely new applications.
However, this approach is not directly applicable to 3D scene understanding, where real-world 3D data lacks web-scale abundance and demands labor-intensive ground truth annotations.

While recent efforts aim to reduce this dependency through self-supervision~\cite{jiang2023selfsupervised-pretraining,zhu2023ponderv2}, distillation~\cite{openscene, yue2024improving}, denoising~\cite{vogel2024p2p}, or open-set scene understanding~\cite{ovseg, wu2024ptv3, yilmaz2024opendas, takmaz2025search3d, engelmann2024opennerf}, state-of-the-art 3D segmentation methods~\cite{wu2024ptv3, nekrasov2021mix3d, mask3d, yue2023agile3d} still rely on some level of direct supervision.
Consequently, annotated data remains essential for learning these tasks, and constructing datasets of comparable scale to those in language and image generation remains a significant challenge. In this paper, we contribute the largest 3D real-world indoor semantic dataset and investigate key questions in 3D scene understanding: Is real-world data preferable to synthetic data? How can labeling efforts be minimized? Do current models benefit from increased real-world data?

\input{table_and_figure_tex/fig_1_teaser.tex}

To address these questions, we leverage ARKit-Scenes~\cite{baruch2021arkitscenes}, a large-scale 3D indoor dataset consisting of 3D reconstructions and RGB-D frames captured with consumer tablets. Although these scenes are annotated with 3D object bounding boxes, they lack the per-point annotations necessary for training competitive 3D segmentation models. To overcome this limitation, we augment the dataset with per-point semantic labels created through an automated pipeline. This approach enables us to produce the significantly larger dataset compared to prior 3D semantic segmentation dataset (see Fig.~\ref{fig:datasets-scale}). Our dataset is suitable for (pre-)training any 3D semantic segmentation model. To validate the effectiveness of these extensive yet imperfect annotations, we use them to re-train various models and conduct comprehensive evaluations on widely-used 3D semantic segmentation benchmarks~\cite{scannet200,dai2017scannet}.

More specifically, we build on top of the recent LabelMaker~\cite{labelmaker} pipeline, which we extend into LabelMakerV2 with more and updated base models, a more general input data structure, as well as deployment scripts for large clusters through docker or SLURM.
Using this pipeline, we process the entire ARKitScenes dataset, which takes 48'000 GPU hours on Nvidia 3090 GPUs.
We further scale our pipeline beyond ARKitScenes to arbitrary scenes by integrating the iOS app Scanner 3D into LabelMakerV2, enabling automatic annotation of scenes recorded with consumer iPhones.
In experiments, we use our automatically generated ARKitScenes labels to pre-train the currently most-used 3D segmentation methods, MinkowskiNet~\cite{choy20194d} and PTv3~\cite{wu2024ptv3}.
We find that without labeling any data manually,
extending the scale of real-world training data improves the performance of both models on multiple benchmarks, or achieves the same performance as with even more synthetic training data.

In summary, we answer the following research question: \emph{``Does large-scale pre-training with automatic labels show similar trends in 3D as it does for language and image tasks?"} through the following key contributions:
\begin{itemize}[leftmargin=*]
    \item Generating the largest existing real-world 3D dataset with dense semantic annotations on 186 classes.
    \item Improving over state-of-the-art PointTransformer on ScanNet200 by 2.1\%, on tail classes by 5.5\% mIoU.
    \item Trade-off analysis between training in unsupervised settings, on synthetic data, and on auto-labeled real-world data providing guidance for future data scaling efforts.
\end{itemize}

%% file: table_and_figure_tex/fig_1_teaser.tex
\begin{figure}
    \centering
    \resizebox{\linewidth}{!}{
        \begin{tikzpicture}

            \definecolor{traincolor}{RGB}{135, 180, 235}
            \definecolor{valcolor}{RGB}{214, 183, 57}
            \definecolor{testcolor}{RGB}{144, 180, 74}

            \draw[->] (0,0) -- (12.5,0) node[right]{num scenes};
            \draw[-] (0,0) -- (0,5.9) node[above]{};

            \foreach \x/\label in {0/0, 2/1000, 4/2000, 6/3000, 8/40000, 10/5000, 12/6000} {
                    \draw (\x,0) -- (\x,-0.05) node[below] {\label};
                }
            \foreach \y in {0.75, 2.25, 3.75, 5.25} {
                    \draw (0, \y) -- (-0.1, \y) node[below] {};
                }

            \node[draw, fill=white, anchor=north east] at (14,5.9) {
                \begin{tabular}{l l}
                    \textcolor{traincolor}{\rule{0.07\linewidth}{0.02\linewidth}} & train \\
                    \textcolor{valcolor}{\rule{0.07\linewidth}{0.02\linewidth}}   & val   \\
                    \textcolor{testcolor}{\rule{0.07\linewidth}{0.02\linewidth}}  & test  \\
                \end{tabular}
            };

            \node[align=center, font=\bfseries\large] at (6.75,5.5) {
                Datasets Size\\
            };

            \fill[traincolor] (0,0.25) rectangle (8.942,1.25);
            \fill[valcolor] (8.942,0.25) rectangle (9.49,1.25);
            \fill[testcolor] (9.49,0.25) rectangle (10.038,1.25);
            \draw[black, line width=0.5pt] (0,0.25) rectangle (8.942,1.25);
            \draw[black, line width=0.5pt] (8.942,0.25) rectangle (9.49,1.25);
            \draw[black, line width=0.5pt] (9.49,0.25) rectangle (10.038,1.25);

            \node[align=left, anchor=west] at (10.038,0.75) {\textbf{ARKit LabelMaker}\\ \textbf{(186 classes)}};

            \fill[traincolor] (0,1.75) rectangle (2.402,2.75);
            \fill[valcolor] (2.402,1.75) rectangle (3.026,2.75);
            \fill[testcolor] (3.026,1.75) rectangle (3.226,2.75);

            \draw[black, line width=0.5pt] (0,1.75) rectangle (2.402,2.75);
            \draw[black, line width=0.5pt] (2.402,1.75) rectangle (3.026,2.75);
            \draw[black, line width=0.5pt] (3.026,1.75) rectangle (3.226,2.75);

            \node[align=left, anchor=west] at (3.226,2.25) {ScanNet/ScanNet200~\cite{dai2017scannet,scannet200}\\ (20/200 classes)};

            \fill[traincolor] (0,3.25) rectangle (0.812,4.25);
            \draw[black, line width=0.5pt] (0,3.25) rectangle (0.812,4.25);
            \node[align=left, anchor=west] at (0.812,3.75) {S3DIS~\cite{s3dis}\\ (13 classes)};

            \fill[traincolor] (0,4.75) rectangle (0.46,5.75);
            \fill[valcolor] (0.46,4.75) rectangle (0.56,5.75);
            \fill[testcolor] (0.56,4.75) rectangle (0.66,5.75);
            \draw[black, line width=0.5pt] (0,4.75) rectangle (0.46,5.75);
            \draw[black, line width=0.5pt] (0.46,4.75) rectangle (0.56,5.75);
            \draw[black, line width=0.5pt] (0.56,4.75) rectangle (0.66,5.75);
            \node[align=left, anchor=west] at (0.66,5.25) {ScanNet++~\cite{yeshwanthliu2023scannetpp}\\ (100 classes)};

        \end{tikzpicture}
    }
    \caption{\label{fig:datasets-scale}Our LabelMaker annotation data creates the world's largest real-world 3D scene annotation dataset.}
\end{figure}

%% file: sec/2_related_works.tex
\section{Related Works}
\label{sec:related_works}

\paragraph{Datasets for 3D semantic segmentation.}
3D semantic segmentation classifies each point in a 3D point cloud into a set of predefined semantic categories.
Prominent datasets for training and evaluation include ScanNet~\cite{dai2017scannet}/ScanNet200~\cite{scannet200} consisting of 1.5k scenes, and the Stanford 3D Indoor Scene Dataset~\cite{s3dis} (S3DIS), which comprises 6 large-scale indoor areas with 271 rooms.
Both datasets include RGB-D frames captured in the real world. In addition to real-world datasets, Structured3D~\cite{Structured3D} is a photo-realistic synthetic dataset with 3.5K house designs, and Replica~\cite{replica} provides 18 high-quality reconstructed scenes.
ARKitScenes~\cite{baruch2021arkitscenes} is the most extensive collection of indoor scenes to date, with 5047 scans of 1661 unique scenes. RGB-D data is recorded with an Apple iPad equipped with a built-in LiDAR scanner.
High-quality surface reconstruction and the bounding box for object detection are also provided.
However, per-point annotations for semantic or instance segmentation are not included, so it cannot be directly used for training 3D segmentation models.

\paragraph{LabelMaker.}~\citet{labelmaker} is an automatic 3D semantic segmentation annotation pipeline that consolidates outputs from state-of-the-art 2D and 3D segmentation models with an additional feature for translating frame-wise 2D labels into consistent 3D point cloud labels.
In this work, we employ an enhanced version of LabelMaker to create 3D semantic segmentation annotations for ARKitScenes.

\paragraph{3D semantic segmentation models.}
Deep learning models for processing 3D input data can be classified into three main categories: voxel-based, point-based, and transformer-based methods.
Voxel-based methods transform points into fixed-sized voxel grids before processing them, such as the popular MinkowskiNet~\cite{choy20194d}.
Mix3D~\cite{nekrasov2021mix3d} enhances MinkowskiNet through effective 3D data augmentation techniques.
PonderV2~\cite{zhu2023ponderv2} explores self-supervised learning from RGB-D data to improve the performance of the MinkowskiNet architecture.
Point-based methods includes \cite{pointnet,pointnet++,pointcnn,superpoint,spidercnn,kpconv,pointwisecnn,pcnn_arxiv}.
However, there is a recent shift from models based on point-wise convolutions to point-based transformer models~\cite{transformer1,transformer2,transformer3,mask3d}.
Notable examples include PointTransformer~\cite{transformer3} and its successors PTv2~\cite{wu2022ptv2}, and PTv3~\cite{wu2024ptv3}, which are developed towards better efficiency and scalability.
Point\,Prompt\,Training~\cite{wu2024ppt}(PPT) introduces a novel training paradigm enabling the simultaneous training of multiple datasets with diverse label spaces.
Combining PTv3 with PPT achieves state-of-the-art performance on the ScanNet/ScanNet200 semantic segmentation benchmark.

In this paper, we address a key limitation of existing datasets for 3D semantic segmentation: their small scale. We hypothesize that this constraint hampers the performance of commonly used models.

%% file: sec/3_method.tex
\section{Method}
\label{sec:method}

\subsection{LabelMaker Revisted}

\input{table_and_figure_tex/fig_2_labelmaker_pipeline.tex}

As we build on LabelMaker~\cite{labelmaker}, we briefly review its key steps. LabelMaker is an automatic pipeline for 2D and 3D semantic annotation, producing labels comparable in quality to human annotations~\cite{dai2017scannet}.
It automatically generates semantic labels by leveraging an ensemble of base models to predict pixel-level semantics for each frame in an RGB-D trajectory.
Since the base models predict segmentations in different label spaces (based on their training data), their predicted semantic labels are then mapped to a unified label space.
Only through this mapping, the different base models can be used in a subsequent ensemble.
Thus, \cite{labelmaker} defined a mapping from every label space into a carefully curated label space based on \emph{wordnet synkeys}~\cite{wordnet}.
After mapping all base model predictions to the unified label space, they are aggregated into a single consensus per frame of the RGB-D trajectory.
This is the first stage of per-frame denoising.
As the RGB-D trajectory provides multi-view information of the scene, the individual frames can be further denoised by lifting 2D predictions onto 3D points and performing per-point voting.
The final labels can either be directly used as 3D labels for 3D semantic segmentation or projected into 2D and be used for training or evaluating 2D semantic segmentation models.
In this paper, we improve this pipeline to robustly scale to large-scale datasets and show its benefit for pretraining 3D semantic segmentation models.
In the following, we describe the improvements in more detail.

\subsection{Improving LabelMaker for Scaling}
While LabelMaker~\cite{labelmaker} introduced an automatic labeling tool that produces annotations comparable to human annotators, we enhance the pipeline with two modifications to further improve its performance, ensuring robust high-quality annotation for large-scale datasets. The complete pipeline is shown in Figure~\ref{fig:pipeline-task-dependency}.

\paragraph{Integrating Grounded-SAM.}
LabelMaker~\cite{labelmaker} employs several state-of-the-art base models in its ensemble, but does not utilize Segment Anything (SAM)~\cite{kirillov2023segment}, a 2D segmentation model trained on large-scale datasets that generalizes robustly across diverse scenarios.
To scale LabelMaker to any environment, we aim to integrate this prior into the pipeline. However, efficiently leveraging this model for semantic segmentation is not straightforward. To address this, Grounded SAM combines Grounding DINO~\cite{liu2023grounding} with SAM~\cite{kirillov2023segment}.
Grounding DINO predicts instance bounding boxes based on semantic labels or natural language, while SAM generates high-quality segmentation masks for these boxes. We integrate this model by adapting it to LabelMaker's unified label space, allowing it to contribute as an additional vote in the ensemble.

\input{table_and_figure_tex/fig_3_g_align_qualitative.tex}

\paragraph{Aligning to Gravity.}
For optimal performance, many semantic segmentation models require the gravity direction to be aligned with the coordinate system used during training. However, large-scale datasets are not inherently gravity-aligned. For instance, in ARKitScenes, occasional phone rotations introduce inconsistencies in the orientation of 2D images.
Passing these misoriented images to LabelMaker's base models degrades performance and leads to misclassifications, such as confusing the floor with the ceiling (see Fig.~\ref{fig:g_align_qualitative}).
Therefore, we project the sky direction, corresponding to the z-axis of ARKit’s pose coordinate system (derived from the IMU), onto each 2D frame. We then compute the angle $\alpha$ between the sky direction and the upward direction.
Given this angle, we rotate the image by $k\cdot \frac{\pi}{2}$, where $k = \arg \min_{s}(|s\frac{\pi}{2} - \alpha|)$ to align the sky direction roughly upward, and rotate the predicted segmentation back to its original orientation after inference to align it with its coordinate system.

\paragraph{Omission of NeuS.}
In LabelMaker~\cite{labelmaker}, an implicit surface model, NeuS~\cite{wang2021neus} with a semantic head, is trained per scene as an optional 3D lifting and 2D denoising step. We omit this step due to its high computational cost. Moreover, NeuS optimizes its own scene geometry, which is poorly constrained in ‘inside-out’ scans of ARKitScenes, making point cloud label generation more complex than a simple coordinate lookup. Instead, we retain per-point voting from \cite{labelmaker}, as it proved the most stable in our initial exploration.

%% file: table_and_figure_tex/fig_2_labelmaker_pipeline.tex
\begin{figure}[t!]
    \centering
    \includegraphics[width=0.85\linewidth]{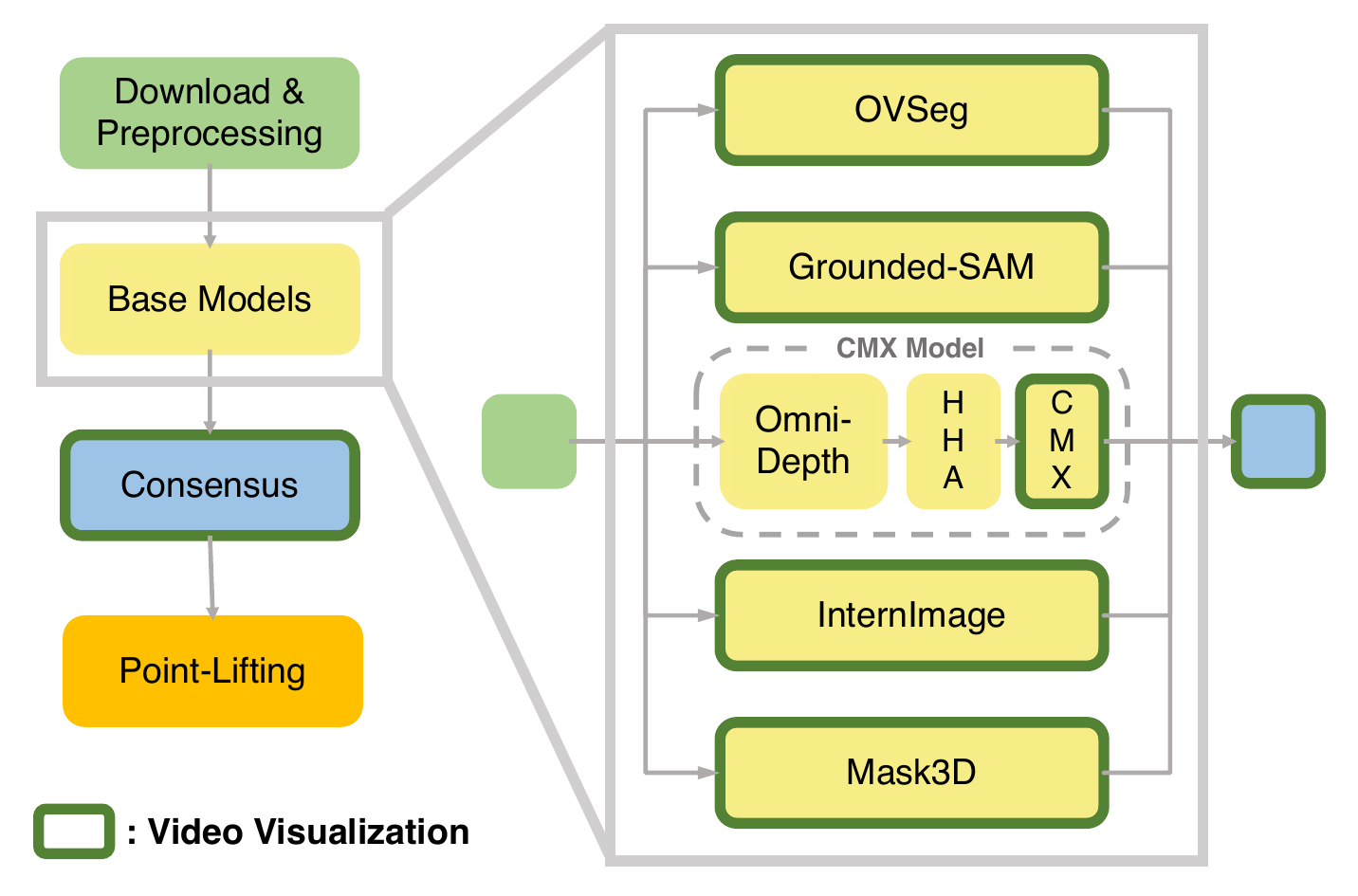}
    \caption{\label{fig:pipeline-task-dependency}\textbf{Dependency graph of the LabelMakerV2 pipeline}.
        Our LabelMakerV2 pipeline has a clear dependency structure that has to be handled in the distributed processing of the data.
        This has to be especially respected when recovering from job failure.
        There, our recovery strategy checks for unfinished jobs in the dependency graph before submitting any new jobs to avoid unnecessarily wasting compute resources.
        The boxes with thick green frame donotes visualizable tasks.
        These are used during inspection and job quality assurance.}
\end{figure}

%% file: table_and_figure_tex/fig_3_g_align_qualitative.tex
\begin{figure}[t!]
    \centering
    \includegraphics[width=0.49\linewidth]{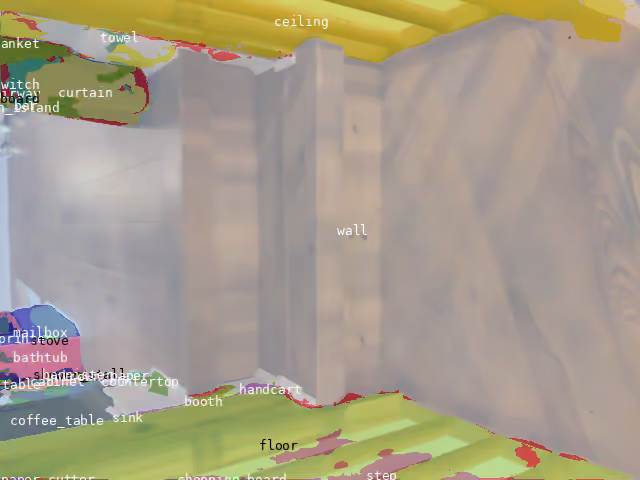} %
    \includegraphics[width=0.49\linewidth]{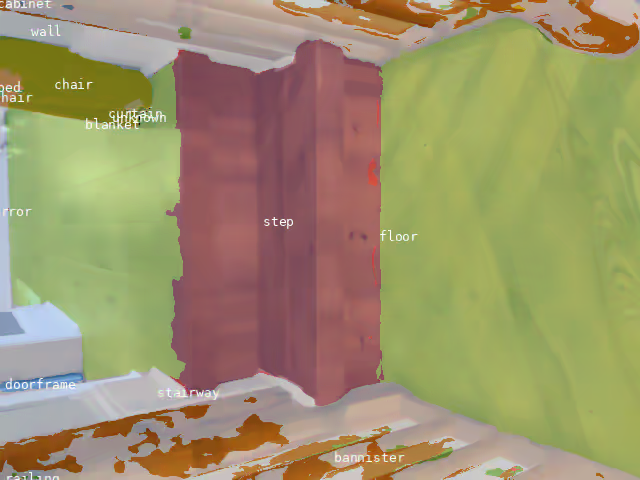}\\
    \includegraphics[width=0.49\linewidth]{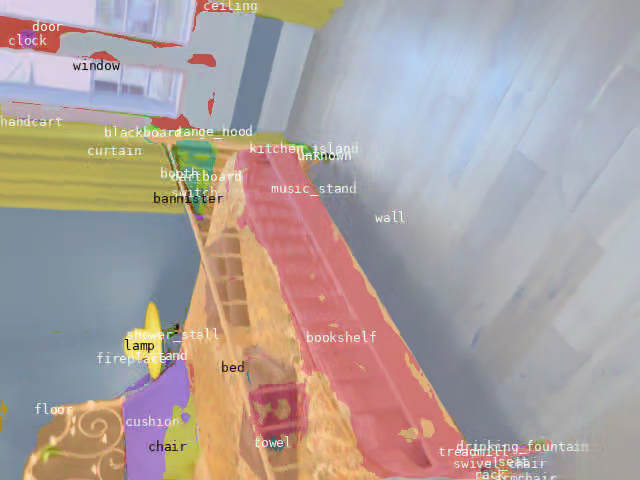} %
    \includegraphics[width=0.49\linewidth]{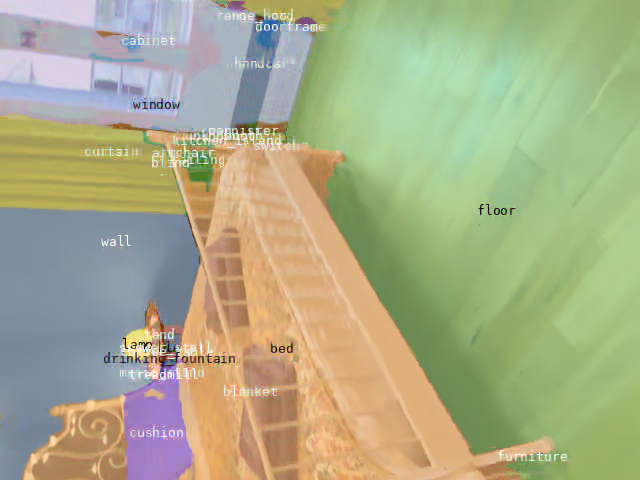}\\
    \includegraphics[width=0.49\linewidth]{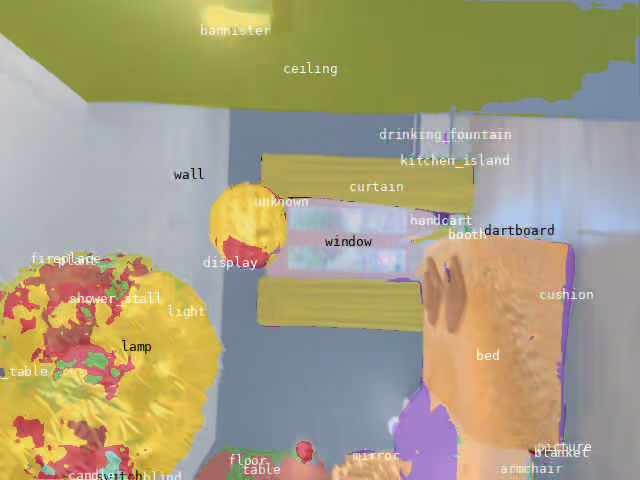} %
    \includegraphics[width=0.49\linewidth]{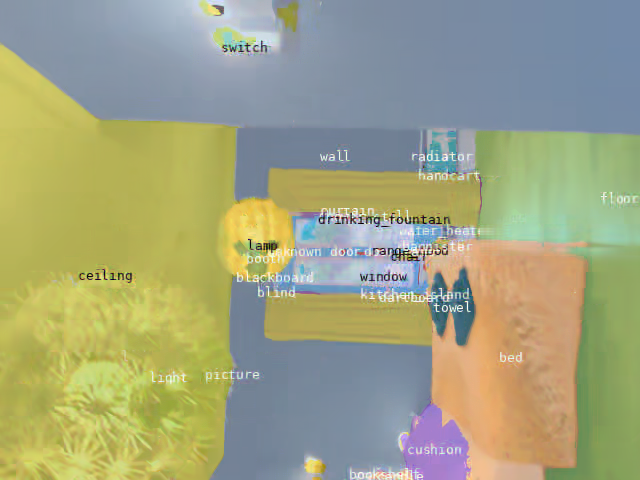}\\
    {\small \textit{Without gravity alignment}
    \hspace{15px}
    \textit{With gravity alignment}}
    \caption{\textbf{Qualitative Evaluation of Gravity Alignment.}
        LabelMaker annotation with and without gravity alignment.
        Without gravity alignment, floors may be misclassified as walls, walls as ceilings, as well as other orientation-dependent objects.}
    \label{fig:g_align_qualitative}
\end{figure}

%% file: sec/4_results.tex
\section{Results}
\label{sec:results}

\subsection{Baselines}
We evaluate the effectiveness of our ARKitScenes LabelMaker dataset using two well-established and distinct network architectures: MinkowskiNet~\cite{choy20194d} and PointTransformer~\cite{pointcept2023,wu2022ptv2,wu2024ppt,wu2024ptv3}.
MinkowskiNet remains the foundation of many top-performing models in 3D semantic segmentation benchmarks, with several modifications~\cite{nekrasov2021mix3d,zhu2023ponderv2} proposed to enhance its performance.
PointTransformer~\cite{wu2024ptv3}, a more recent architecture, achieves state-of-the-art results on the ScanNet and ScanNet200 benchmarks. Given that transformers generally benefit from large-scale training data, we also train on this architecture. From these two architectures, we derive three relevant baselines:

\paragraph{Vanilla MinkowskiNet.}
This is the standard MinkowskiNet model based on~\cite{choy20194d}, which most 3D semantic segmentation methods compare to.
In this paper, we use the commonly used `Res16UNet34C`variant of MinkowskiNet to guarantee fair comparison to all other baselines.

\paragraph{Mix3D~\cite{nekrasov2021mix3d}} is a data augmentation method for large-scale 3D scene segmentation that creates new training samples by merging two augmented scenes, effectively placing object instances into novel, out-of-context environments. This approach encourages models to infer semantics from local structures rather than relying on overall scene context. MinkowskiNet shows significant performance improvements when trained with Mix3D.

\paragraph{PonderV2~\cite{zhu2023ponderv2}.} Addressing the scarcity of 3D annotations involves two strategies, unsupervised feature learning and automated pseudo-labeling.  PonderV2~\cite{zhu2023ponderv2} represents the former, using neural rendering objectives to learn 3D features without semantic annotations. Our work explores the latter, generating large-scale pseudo-labels through automatic annotation. While PonderV2 currently achieves state-of-the-art results in unsupervised settings, we compare both paradigms to quantify the relative merits of label-free feature learning versus pseudo-label-driven supervision for scaling 3D segmentation models.

\paragraph{PointTransformerV3 (PTv3)~\cite{wu2024ptv3}}
is a recently proposed method to accelerate transformer architectures and enable large-scale training by jointly training multiple datasets with diverse label spaces.
This contrasts with the LabelMaker~\cite{labelmaker} approach, which translates label spaces to a common one before training.
The combination of PTv3 and PPT achieves state-of-the-art performance on the ScanNet/ScanNet200 semantic segmentation benchmarks.

\subsection{Datasets and Evaluation Metrics}

\input{table_and_figure_tex/tab_1_datasets_scale.tex}

\paragraph{ScanNet~\cite{dai2017scannet}} comprises 1513 densely annotated scans across 707 distinct indoor scenes,
totaling 2.5 million RGB-D frames.
It stands as one of the most widely used and influential benchmark datasets for indoor 3D scene understanding.
It is annotated by humans using the NYU40 label space and evaluated on a subset of 20 classes from NYU40.

\paragraph{ScanNet200~\cite{scannet200}.}
While only 20 classes are used in the ScanNet benchmark, the original dataset is annotated with many more classes.
ScanNet200~\cite{scannet200} leverages these annotations and organizes them into a new benchmark with 200 classes that are of higher-resolution than the original ScanNet classes.
Given the large-number of different categories generated by our LabelMakerV2 pipeline,
we also pre-train the models for this task and evaluate them on the ScanNet200 benchmark.

\paragraph{ScanNet++~\cite{yeshwanthliu2023scannetpp}} is a dataset of 460 high-resolution 3D indoor scenes with dense semantic and instance annotations, captured using a high-precision laser scanner and registered images from a DSLR camera and RGB-D streams.
It focuses on long-tail and multi-labeled annotations.
Models are typically evaluated on 100 classes.

\paragraph{Structured3D~\cite{Structured3D}} is a large-scale indoor synthetic RGB-D dataset featuring 6519 training scenes and 1697 test scenes.
It is annotated with a label space of 25 classes.
Structured3D is only used in PTv3+PPT joint training and we adopt pre-processed version of these two datasets from \cite{wu2024ptv3}.

\paragraph{Matterport3D~\cite{Matterport3D}} is a large-scale RGB-D dataset containing 10,800 panoramic views from 194,400 RGB-D images across 90 building-scale scenes.
We map the ScanNet200 label space to Matterport3D and use this dataset for zero-shot evaluation of our trained model.

\paragraph{ARKit LabelMaker (Ours).}
This is the dataset generated with our method described above. The resulting dataset contains 5019 scenes, from which we take 4471 for training and 548 for validation according to the official train-val split provided by the original ARKitScenes~\cite{baruch2021arkitscenes} dataset.
For every scene, we created 3D point cloud associated per-point semantic labels in the original LabelMaker wordnet label space (186 classes). In some experiments, we project the dataset from wordnet label space to ScanNet200 label space to train it together with ScanNet200 dataset. We denote this converted dataset as {ARKit LabelMaker\textsuperscript{SN200}}.
To increase efficiency and make the experimental settings comparable to previous studies, we perform down-sampling on 3D meshes to a voxel size of 2 cm.
Normal information is preserved and down-sampled simultaneously.
\Cref{tab:dataset-scale} illustrates the scale of each dataset.
ARKit LabelMaker dataset is the largest annotated real-world indoor semantic dataset.

\input{table_and_figure_tex/tab_2_scannet200.tex}

\input{table_and_figure_tex/tab_3_scannet.tex}

\paragraph{Metrics.}
We follow the standard metrics of the ScanNet 3D semantic segmentation task and compute the mean and per-class intersection-over-union (IoU), the mean per-class accuracy (mAcc) and the total per-point accuracy (tAcc).

\subsection{Experiment Settings}
We adopt three approaches to evaluate the effectiveness of our ARKitScenes LabelMaker dataset.

\paragraph{Pre-training.}
To investigate whether automatic labels are useful to learn strong features from imperfect annotations, we pre-train both, MinkowskiNet and PointTransformerV3, on our generated ARKit LabelMaker\textsuperscript{SN200} dataset.
Afterwards, we fine-tune the pretrained models on the ScanNet and ScanNet200 dataset, respectively.

For MinkowskiNet, we employ the Res16UNet34C architecture as our backbone model.
During pre-training, we utilize the AdamW optimizer with a learning rate of 0.01 and OneCycleLR scheduler, training the network for 600 epochs.
If the label space is changed for fine-tuning, we replace the classification head and exclusively train it with the same learning rate setting until convergence while the rest of the model is fixed.
Then, the entire network undergoes fine-tuning with a learning rate of 0.001, while other settings are kept unchanged.

For PTv3~\cite{wu2024ptv3}, we follow \cite{wu2024ptv3} employing the AdamW optimizer with OneCycleLR for 800 epochs of training.
Similar to the fine-tuning of MinkowskiNet, we initially freeze the backbone and then solely train the classification head until convergence.
Then, we fine-tune on ScanNet or ScanNet200 with a reduced learning rate of 0.0006.
Besides ARKit LabelMaker\textsuperscript{SN200}, we also pre-train PTv3 with ARKit LabelMaker in wordnet label space as the mapping from wordnet to ScanNet200 may reduce class diversity.

\paragraph{Co-training with ScanNet200.}
With this experiment, we investigate whether ARKit LabelMaker can be seamlessly combined with existing datasets to increase dataset size and improve model performance. To this end, we merge ARKit LabelMaker\textsuperscript{SN200} with ScanNet200 and train a MinkowskiNet from scratch. Due to resource constraints, we conduct this experiment only with MinkowskiNet. The training setup follows the exact pre-training procedure described earlier for MinkowskiNet.

\paragraph{Joint-training.}
We employ PTv3+PPT for joint training across multiple datasets and label spaces. In addition to ScanNet/ScanNet200, ScanNet++, and Structured3D, we incorporate our ARKit LabelMaker dataset. To maximize semantic class exposure, we adopt LabelMaker's WordNet label space. Our training setup follows the exact PTv3+PPT configuration from \cite{wu2024ptv3}, using the AdamW optimizer with a OneCycleLR scheduler and a learning rate of 0.05. We also integrate the LabelMaker WordNet label space into the normalization layer and final classification head.

\subsection{Experiments}

\input{table_and_figure_tex/tab_4_alc_percentage_scaling.tex}

\input{table_and_figure_tex/tab_5_ablation_gsam.tex}

In \Cref{tab:scannet-results}, we present the results for the ScanNet dataset.
For MinkowskiNet~\cite{choy20194d}, we show that pre-training on our large-scale, real-world ARKit LabelMaker\textsuperscript{SN200} dataset not only significantly improves the mean intersection-over-union compared to vanilla training but also outperforms other pre-training variants.
Similar trends are observed in \Cref{tab:scannet200-results}, which reports results on the ScanNet200 dataset. Co-training MinkowskiNet on ScanNet200 further confirms that our ARKit LabelMaker\textsuperscript{SN200} dataset enhances training without introducing a domain gap relative to ScanNet200.

\paragraph{Comparison to Self-supervision.}
\Cref{tab:scannet-results} shows that pre-training on our imperfect yet automatically generated labels outperforms self-supervised pre-training (PonderV2~\cite{zhu2023ponderv2}) and extensive data augmentation (Mix3D~\cite{nekrasov2021mix3d}). This highlights the importance of direct supervision with large-scale training data for effective 3D segmentation.

\paragraph{Comparison to Training on Synthetic Data.}
Table~\ref{tab:scannet200-results} shows that PTv3 pre-trained on ARKit LabelMaker achieves comparable or superior performance to large-scale multi-dataset joint training. The approach in~\cite{wu2024ptv3} relies heavily on Structure3D, a synthetically generated dataset, for pretraining, motivating a deeper analysis in Table~\ref{tab:scaling-alc}.
We compare pretraining on different subset sizes of ARKit LabelMaker to an equivalent amount of synthetic data from Structure3D~\cite{Structured3D}. Even at smaller scales, pretraining with real-world ARKit LabelMaker data proves more effective than synthetic data, which can even degrade performance at limited sizes. These findings strongly indicate that auto-labeling real-world recordings is significantly more effective per data point than relying on synthetic data.

\paragraph{Effect on Long-tail Classes.}
For Point Transformer, integrating our ARKit LabelMaker into the joint training of PTv3+PPT results in a noticeable boost in validation and test mIoU. This version of PTv3 achieves state-of-the-art performance on both ScanNet and ScanNet200, with the latter benefiting significantly from improved tail class mIoU (\Cref{tab:scannet200-results}, \Cref{fig:tail_class_spider_plot}).
When comparing fine-tuned and vanilla models on the ScanNet200 validation set, our trained model shows a performance gain of +0.5\% on head classes and +2.4\% on tail classes. Compared to PTv3-PPT trained without ARKit LabelMaker, our model achieves a +0.8\% gain on head classes and +5.5\% on tail classes on the ScanNet200 test set.
This effect persists even in a zero-shot setting on the Matterport3D dataset (\Cref{tab:matterport3d-ood}). Including ARKit LabelMaker in training yields similar performance gains across Top-40 to Top-160 class sets.
For MinkowskiNet, we do not observe a consistent improvement in tail class performance.
On ScanNet200, validation mIoU slightly declines.
In zero-shot evaluation on Matterport3D, the performance gain from adding ARKit LabelMaker\textsuperscript{SN200} over MinkowskiNet trained solely on ScanNet200 diminishes from Top-40 to Top-160 class sets.

\input{table_and_figure_tex/tab_6_ablation_g_align.tex}

\input{table_and_figure_tex/fig_4_tail_class_spider_plot.tex}

\subsection{Ablation Studies}

\paragraph{Different Training Regimes.}
Between pre-training with ARKit LabelMaker and co-training with aligned labels, we observe no significant difference in \Cref{tab:scannet200-results}. Co-training is primarily relevant for MinkowskiNet, as PTv3 allows for joint training across different output label spaces through PPT~\cite{wu2024ptv3}. Increasing the dataset scale further through joint training on multiple datasets generally leads to better performance than pre-training solely on ARKit LabelMaker.

\paragraph{Evaluation of LabelMakerV2.}
In \Cref{tab:eval_scannet}, we evaluate our updated LabelMakerV2 pipeline on five ScanNet scenes with five manual annotations in LabelMaker's WordNet label space, sourced from the original LabelMaker. The current pipeline achieves higher accuracy than even ScanNet's original annotations.
In \Cref{tab:eval_arkit}, we conduct an ablation study to assess the performance improvements from integrating Grounded-SAM and gravity alignment. For this study, we manually annotate three ARKitScenes and provide additional visualizations in \Cref{reb:fig:qualitative_labels}. \Cref{fig:g_align_qualitative} qualitatively demonstrates how gravity alignment corrects prediction errors in a video with continuous viewpoint rotation.

\input{table_and_figure_tex/fig_5_scene_viz.tex}

\input{table_and_figure_tex/fig_5_mobile_capture_viz.tex}

\input{table_and_figure_tex/tab_7_matterport3d_zero_shot.tex}

\input{table_and_figure_tex/tab_8_instance_seg.tex}

\paragraph{Scaling Potential of LabelMaker and PTv3.}
In \Cref{tab:scaling-alc}, we conduct an ablation study by varying the number of pre-training samples from the ARKit LabelMaker dataset and evaluating fine-tuning performance on ScanNet200.
The log-linear relationship between loss and data size is well-documented in large language models.
In our experiments we cannot find such trend for validation loss at either the pre-training or fine-tuning stage.
Instead, we find a log-linear relation between validation mIoU and dataset size from 20\% onwards, which in turn indicates further scaling opportunities for even larger datasets (
\makeatletter
\@ifpackagewith{cvpr}{pagenumbers}{\Cref{reb:fig:scaling}}{Supplementary Fig~E3}
\makeatother
).
We also run a pre-training experiment with the same data volume from the synthetic Structured3D dataset, which showed no performance gain. This further highlights the importance of our pipeline.

\subsection{Transferability to other Domains}

\paragraph{Efficacy in Downstream task.}
Since ARKit LabelMaker is auto-labeled for 3D semantic segmentation, an important question is how useful this data generation approach is for other related 3D perception tasks. To assess this, we evaluate the general features learned through supervised training on our dataset by using PTv3 as the backbone and fine-tuning PointGroup~\cite{pointgroup} for 3D instance segmentation on ScanNet and ScanNet200.
Results in \Cref{tab:instance-seg} show a substantial performance improvement on ScanNet200 when incorporating our ARKit LabelMaker dataset. This suggests that auto-labeling as a method for scaling data is not limited to a single task but can serve as a general pretraining objective for learning robust features, akin to how ImageNet pretraining has become a standard for image encoders.

\paragraph{Zero-shot Evaluation on Matterport3D.}
We perform inference on the Matterport3D~\cite{Matterport3D} dataset using the ScanNet200 label space, and map the labels to the Top-40/80/160 NYU classes as done in \cite{openscene}. PTv3 trained with ARKit LabelMaker surpasses both directly supervised models and OpenScene~\cite{openscene}, an open-vocabulary model. These results indicate that ARKit LabelMaker has sufficient scale and diversity of real-world appearance to strongly improve model generalization.

\paragraph{Data Scaling via Mobile Integration.}
While LabelMaker demonstrates success in large-scale 3D training on ARKitScenes, the dataset's diversity remains limited. To scale our pipeline to arbitrary real-world environments, we integrate the iOS app Scanner 3D into LabelMaker, leveraging modern mobile devices’ ubiquitous RGB-D capture capabilities. This integration enables automatic annotation of scenes recorded with consumer iPhones. We validate scalability by processing two self-captured scenes, a kitchen and a fireplace, recorded using an iPhone 12 Max in a holiday cottage. \Cref{fig:viz} showcases reconstructed scenes and their semantic segmentations, confirming the pipeline’s effectiveness for diverse real-world settings.

\subsection{Limitations \& Broader Impact}
While we enhance LabelMaker~\cite{labelmaker} with an improved point cloud pipeline, we omit the generation of 2D segmentation maps. The computational cost of NeRF-based lifting across the entire ARKitScenes dataset exceeds our available resources, requiring approximately 12 additional GPU hours per scene. An interesting future research direction would be to explore more efficient 2D lifting methods and assess whether training 2D models on this data yields performance gains comparable to those observed for 3D models.
Furthermore, 20 ARKitScenes scenes are excluded from processing due to missing pose data. Since LabelMakerV2 relies on accurate poses, future iterations of the software stack could incorporate techniques to reconstruct missing pose information, enabling broader dataset coverage.
Like the original LabelMaker~\cite{labelmaker}, our improved pipeline does not achieve perfect accuracy. While \cite{labelmaker} demonstrated that its annotations are comparable to crowd-sourced human labels, training on noisy data always carries the risk of introducing systematic errors. For safety-critical applications, rigorous evaluation on accurately annotated data remains essential when leveraging tools like ours for training data generation.
Does large-scale pretraining with automatic labels exhibit similar trends as seen in language and image generation tasks? Our results suggest so, showing measurable improvements across multiple architectures and tasks when (pre)training on ARKit LabelMaker. However, while real-world data proves significantly more effective than synthetic data, the scale of generated 3D data remains orders of magnitude smaller than that of image datasets. Our pipeline facilitates data collection, making it easier to expand training datasets as more scans become available.

%% file: table_and_figure_tex/tab_1_datasets_scale.tex
\begin{table}[t!]
    \centering
    \small
    \resizebox{\linewidth}{!}{
        \begin{tabular}{l| c c c | c | c }
            \toprule
            \textbf{Dataset}          & \#train & \#val & \#test & real        & \#classes \\
            \midrule
            Structured3D              & 6519    & -     & 1697   & \usym{2718} & 25        \\
            \midrule
            S3DIS                     & 406     & -     & -      & \usym{2714} & 13        \\
            ScanNet/ScanNet200        & 1201    & 312   & 100    & \usym{2714} & 20 / 200  \\
            ScanNet++                 & 230     & 50    & 50     & \usym{2714} & 100       \\
            \textbf{ARKit LabelMaker} & 4471    & 274   & 274    & \usym{2714} & 186       \\
            \bottomrule
        \end{tabular}
    }
    \caption{\label{tab:dataset-scale}\textbf{Dataset Size}.
        We provide by far the largest real-world labeled training dataset compared to existing real-world datasets. We provide automatically generated per-point semantic annotations for 4471 training scenes and 548 validation scenes. }
\end{table}

%% file: table_and_figure_tex/tab_2_scannet200.tex
\begin{table*}[!th]
    \centering
    \resizebox{\linewidth}{!}{
        \begin{tabular}{llcccccccc}
            \toprule
            \multirow{2}{4em}{\textbf{Method}} & \multirow{2}{8em}{\textbf{Training Data}}                                   & \multicolumn{4}{c}{\textbf{Val mIoU}} & \multicolumn{4}{c}{\textbf{Test mIoU}}                                                                                                       \\ \cmidrule(lr){3-6} \cmidrule(lr){7-10}
                                               &                                                                             & \textbf{all}                          & \textbf{head}                          & \textbf{ common} & \textbf{tail} & \textbf{all}  & \textbf{head} & \textbf{ common} & \textbf{tail} \\ \midrule
            \multicolumn{10}{c}{MinkUNet~\cite{choy20194d}}                                                                                                                                                                                                                                                         \\ \midrule
            vanilla                            & ScanNet200                                                                  & 27.2                                  & 49.8                                   & 20.8             & \textbf{11.2} & 25.3          & 46.3          & 15.4             & 10.2          \\
            fine-tune (Ours)                   & ARKit LabelMaker\textsuperscript{SN200}  → ScanNet200                       & 28.1                                  & 51.6                                   & 22.3             & 10.6          & \textbf{27.4} & \textbf{49.0} & \textbf{19.4}    & 9.4           \\
            co-training (Ours)                 & ARKit LabelMaker\textsuperscript{SN200}  + ScanNet200                       & \textbf{28.2}                         & \textbf{52.5}                          & \text{22.4}      & 9.8           & -             & -             & -                & -             \\ \midrule
            \multicolumn{10}{c}{PTv3~\cite{wu2024ptv3} }                                                                                                                                                                                                                                                            \\ \midrule
            vanilla                            & ScanNet200                                                                  & 35.2                                  & 56.5                                   & 30.1             & 19.3          & 37.8          & -             & -                & -             \\
            fine-tune (Ours)                   & ARKit LabelMaker\textsuperscript{SN200}  → ScanNet200                       & 36.4                                  & 56.7                                   & 31.2             & 21.6          & -             & -             & -                & -             \\
            fine-tune (Ours)                   & ARKit LabelMaker  → ScanNet200                                              & 37.0                                  & 57.0                                   & 32.6             & \textbf{21.7} & 38.4          & 58.2          & 30.9             & 22.2          \\
            PPT~\cite{wu2024ptv3}              & {\small ScanNet200 + S3DIS + Structure3D → ScanNet200}                      & 36.0                                  & -                                      & -                & -             & 39.3          & 59.2          & \textbf{33.0}    & 21.6          \\
            PPT(Our ablation)                  & {\small ScanNet + ScanNet200 + Structure3D → ScanNet200}                    & 36.3                                  & 56.6                                   & 31.7             & 20.7          & -             & -             & -                & -             \\
            PPT(Ours)                          & {\small{ScanNet + ScanNet200 + ScanNet++ + Structure3D + ARKit LabelMaker}} & \textbf{37.5}                         & \textbf{58.8}                          & \textbf{33.3}    & 20.4          & \textbf{41.4} & \textbf{61.0} & 32.2             & \textbf{27.1} \\
            \bottomrule
        \end{tabular}
    }
    \caption{\label{tab:scannet200-results}\textbf{3D Semantic Segmentation Scores on ScanNet200~\cite{scannet200}.}
        To investigate our large-scale dataset also helps with long-tail categories, we evaluate it on the ScanNet200 dataset.
        For both MinkowskiNet~\cite{choy20194d} and PointTransformerv3~\cite{wu2024ptv3}, we compare it to vanilla training as well as training procedure proposed in \cite{wu2024ptv3}.
        We can show that common neural networks benefit from pre-training on automatically generated large-scale annotations. We also report the head, common and tail classes mean IoU.
    }
\end{table*}

%% file: table_and_figure_tex/tab_3_scannet.tex
\begin{table}[!th]
    \centering
    \resizebox{\linewidth}{!}{
        \begin{tabular}{llcc}
            \toprule
            \textbf{Method}                 & \textbf{Training Data}                                                     & \textbf{val}  & \textbf{test} \\
            \midrule
            \multicolumn{4}{c}{MinkUNet~\cite{choy20194d}}                                                                                               \\ \midrule
            vanilla                         & ScanNet                                                                    & 72.4          & 73.6          \\
            PonderV2~\cite{zhu2023ponderv2} & ScanNet {\scriptsize (self-supervised)} → ScanNet                          & 73.5          & -             \\
            Mix3D~\cite{nekrasov2021mix3d}  & ScanNet                                                                    & 73.6          & 78.1          \\
            fine-tune (Ours)                & ARKit LabelMaker\textsuperscript{SN200}  → ScanNet                         & \textbf{74.1} & -             \\ \midrule
            \multicolumn{4}{c}{PTv3~\cite{wu2024ptv3} }                                                                                                  \\ \midrule
            vanilla                         & ScanNet                                                                    & 77.5          & 77.9          \\
            fine-tune (Ours)                & ARKit LabelMaker\textsuperscript{SN200}   → ScanNet                        & 78.9          & -             \\
            fine-tune (Ours)                & ARKit LabelMaker → ScanNet                                                 & 78.0          & 79.0          \\
            PPT~\cite{wu2024ptv3}           & {\small ScanNet  + S3DIS + Structure3D}                                    & 78.6          & 79.4          \\
            PPT (Ours)                      & {\small{ScanNet+ ScanNet200 + ScanNet++ + Structure3D + ARKit LabelMaker}} & \textbf{79.1} & \textbf{79.8} \\
            \bottomrule
        \end{tabular}
    }
    \caption{\label{tab:scannet-results}
        \textbf{3D Semantic Segmentation Scores}.
        Comparing training strategies for two top-performing models (PointTransformerV3~\cite{wu2024ptv3} and MinkowskiNet~\cite{choy20194d}) on ScanNet20~\cite{dai2017scannet}.
        Adding ARKit LabelMaker\textsuperscript{SN200} through pre-training and co-training improves the performance for both models.
        With PonderV2~\cite{zhu2023ponderv2} and Mix3D~\cite{nekrasov2021mix3d}, we compare large-scale pretraining to two other training strategies.
        Large-scale pre-training is superior to both, extensive data augmentation (Mix3D) and self-supervised pre-training (PonderV2).
    }
\end{table}

%% file: table_and_figure_tex/tab_4_alc_percentage_scaling.tex
\begin{table}[t]
    \centering
    \footnotesize
    \resizebox{\linewidth}{!}{%
        \begin{tabular}{l r}
            \toprule
            \textbf{Pretrain Dataset}                          & \textbf{fine-tune val mIoU} \\ \midrule
            scratch                                            & 32.76                       \\
            10\% ARKit LabelMaker                              & 33.30 (+0.54)               \\
            20\% ARKit LabelMaker                              & 35.29 (+2.53)               \\
            50\% ARKit LabelMaker                              & 36.29 (+3.53)               \\
            100\% ARKit LabelMaker                             & 37.04 (+4.28)               \\  \midrule
            Structured3D (re-sampled to ARKit LabelMaker size) & 32.21                       \\
            \bottomrule
        \end{tabular}
    }
    \caption{\label{tab:scaling-alc} PTv3~\cite{wu2024ptv3} pretrained on different portion of ARKit LabelMaker dataset and Structured3D resampled to be the same size as ARKit LabelMaker. We then fine-tune on ScanNet200 and report validation mIoU.
        These experiments show the scaling potential of PTv3 and the advantage of real-world over synthetic data.
    }
\end{table}

%% file: table_and_figure_tex/tab_5_ablation_gsam.tex
\begin{table}[t]
    \centering
    \small
    \begin{tabular}{lccc} \toprule
        \textbf{Method}\hspace{3.5cm} & \textbf{mIoU} & \textbf{mAcc} & \textbf{tAcc} \\ \midrule
        ScanNet's Annotation          & 17.7          & 21.3          & 70.6          \\
        LabelMakerV1                  & 16.3          & 20.3          & 75.0          \\
        LabelMakerV2 w/o GSAM         & 17.6          & 21.0          & \textbf{77.1} \\
        LabelMakerV2 w/ GSAM          & \textbf{18.5} & \textbf{22.7} & 76.9          \\ \bottomrule
    \end{tabular}
    \caption{\textbf{Evaluation of LabelMaker on 5 ScanNet scenes.}
        We use the same ScanNet scenes in the original LabelMaker's paper with manual annotation in WordNet label space.
        Our LabelMakerV2 improves over ScanNet's manual annotation.}
    \label{tab:eval_scannet}
\end{table}

%% file: table_and_figure_tex/tab_6_ablation_g_align.tex
\begin{table}[t]
    \centering
    \small
    \begin{tabular}{lccc} \toprule
        \textbf{LabelMakerV2 Method Variant} \hspace{0.1cm} & \textbf{mIoU} & \textbf{mAcc} & \textbf{tAcc} \\ \midrule
        w/o GSAM w/o $g$-alignment                          & 7.1           & 10.0          & 45.1          \\
        w/o GSAM \;\;w/ $g$-alignment                       & 10.3          & 13.4          & 70.6          \\
        \;\;w/ GSAM w/o $g$-alignment                       & 7.9           & 11.0          & 44.9          \\
        \;\;w/ GSAM \;\;w/ $g$-alignment                    & \textbf{12.9} & \textbf{15.7} & \textbf{73.6} \\ \bottomrule
    \end{tabular}
    \caption{We annotate three ARKitScenes scenes and perform ablation study of GSAM and gravity alignment. Our results shows that both GSAM and gravity alignment are helpful to the performance. This demonstrate the contribution of our novelty in improving LabelMaker's pipeline. We also provide the visualization of ground truth labels and predictions of these scenes in \Cref{reb:fig:qualitative_labels}.}
    \label{tab:eval_arkit}
\end{table}

%% file: table_and_figure_tex/fig_4_tail_class_spider_plot.tex
\begin{figure}[t]
    \centering
    \includegraphics[width=0.95\linewidth,trim=0px 30px 0 0, clip]{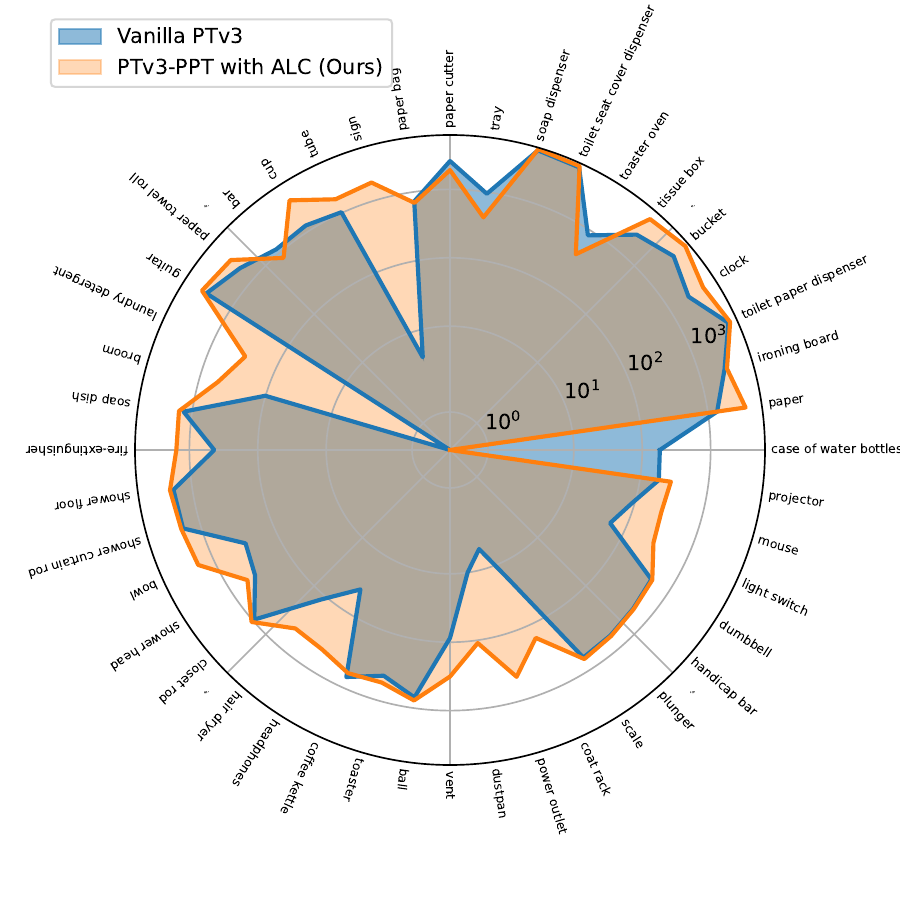}
    \caption{\textbf{Correctly predicted tail class points on ScanNet200 validation set.} We compare the number of correctly predicted points of selected tail class in ScanNet200 validation sets between PTv3 trained from scratch and the PTv3-PPT trained with our datasets. With our dataset, Point Transformer gains more ability to detect rase classes. Tail classes that are not predicted by any models are ignored in this plot, and we present the full tail class performance difference in
        the supplementary.}
    \label{fig:tail_class_spider_plot}
\end{figure}

%% file: table_and_figure_tex/fig_5_scene_viz.tex
\begin{figure}[t!]
    \centering
    \parbox{0.24\linewidth}{\centering \footnotesize \textit{3D Scene}}
    \parbox{0.24\linewidth}{\centering \footnotesize \textit{Manual Annotations}}
    \parbox{0.24\linewidth}{\centering \footnotesize \textit{LabelMaker auto-labels (Ours)}}
    \parbox{0.24\linewidth}{\centering \footnotesize \textit{OpenScene~\cite{openscene}}}\\
    \includegraphics[width=0.24\linewidth]{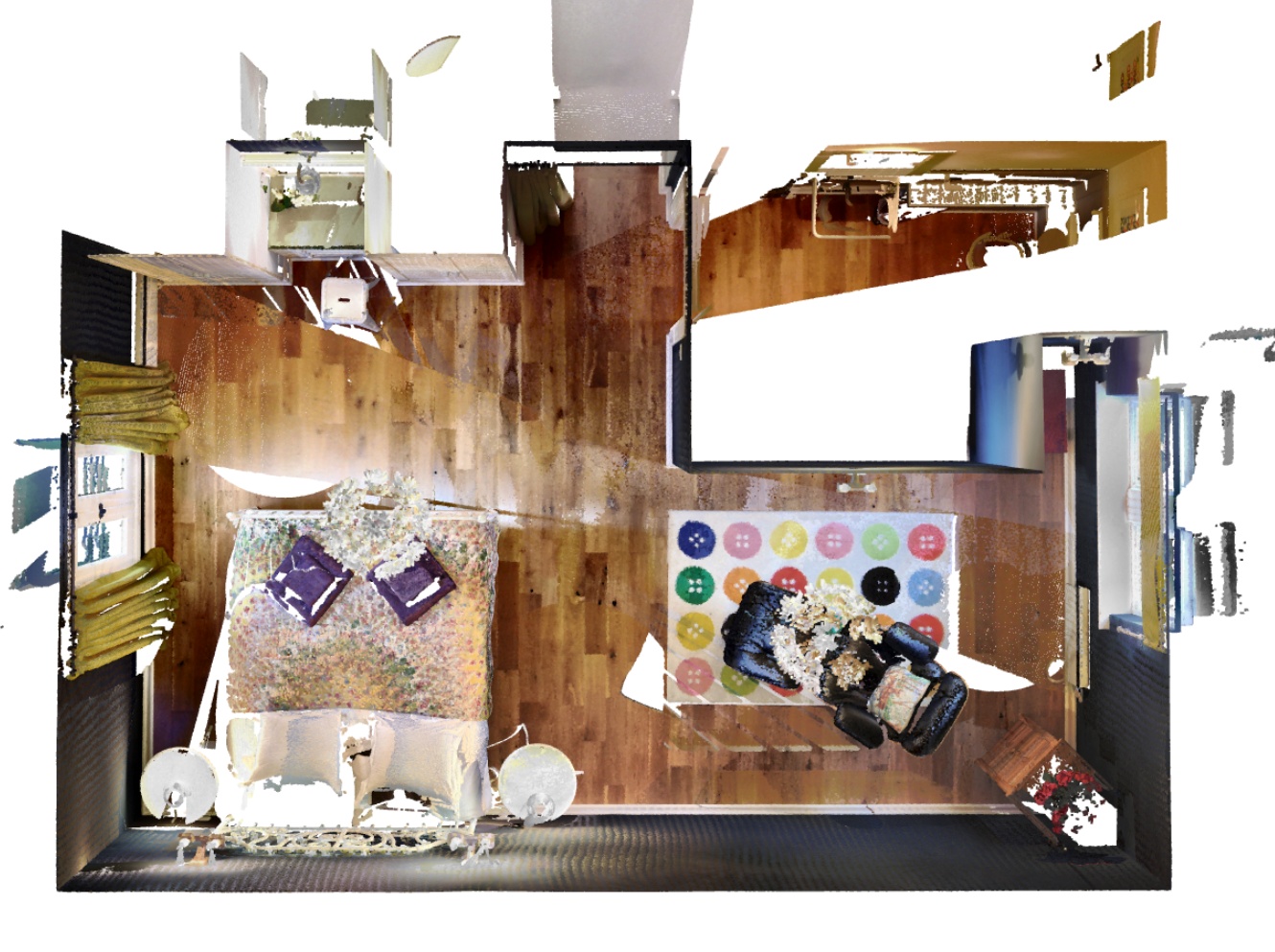}
    \includegraphics[width=0.24\linewidth]{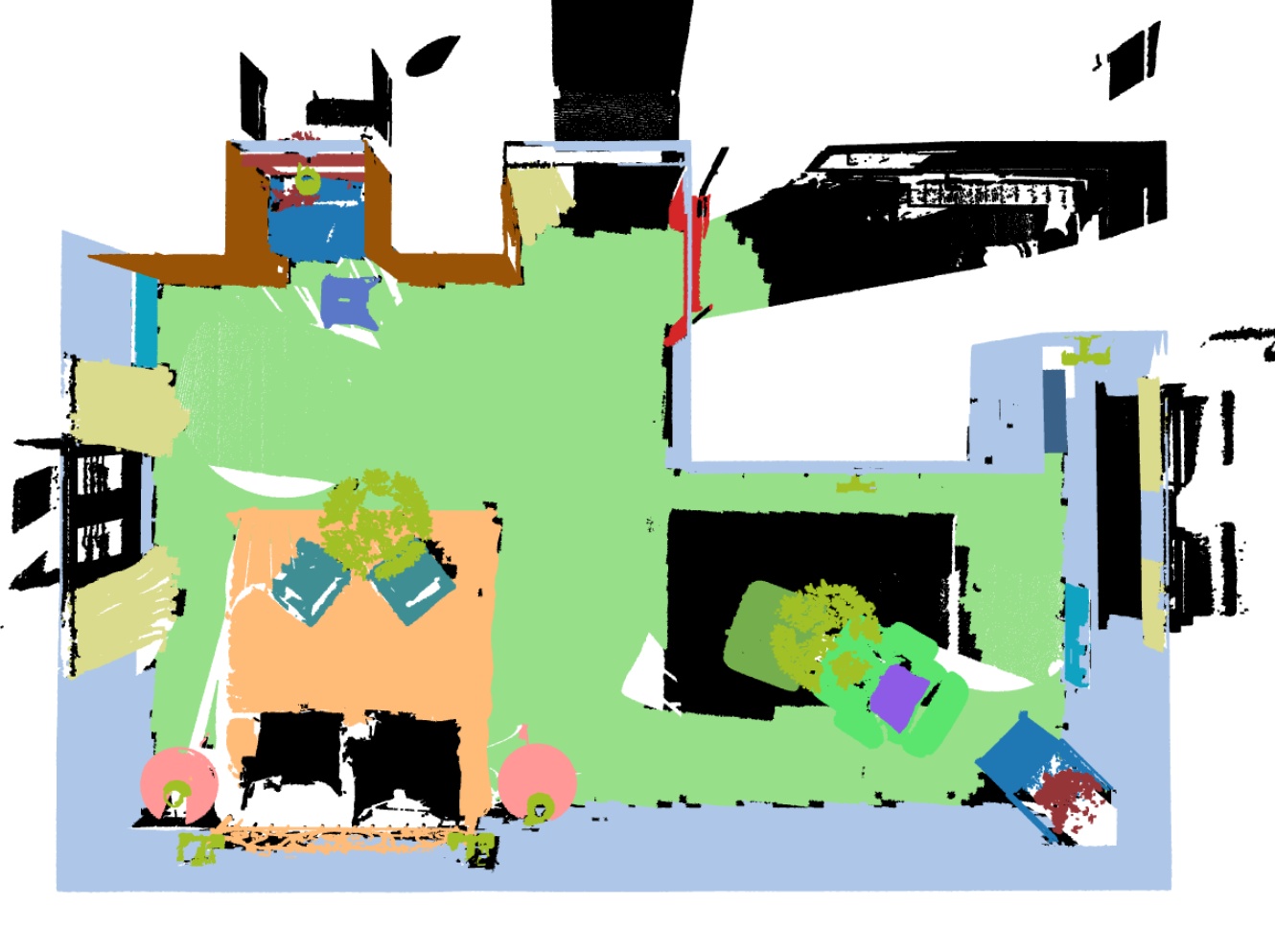}
    \includegraphics[width=0.24\linewidth]{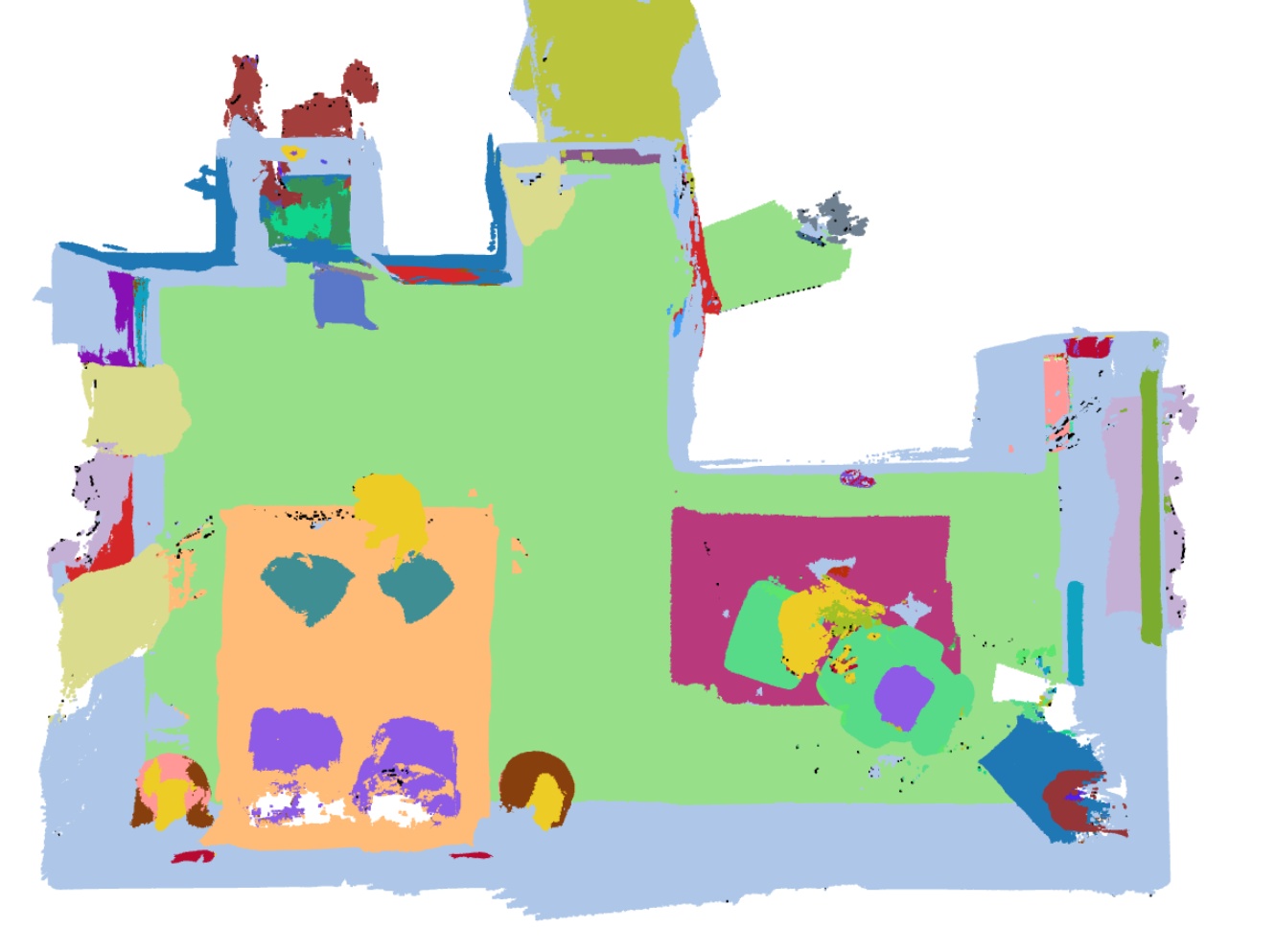}
    \includegraphics[width=0.24\linewidth]{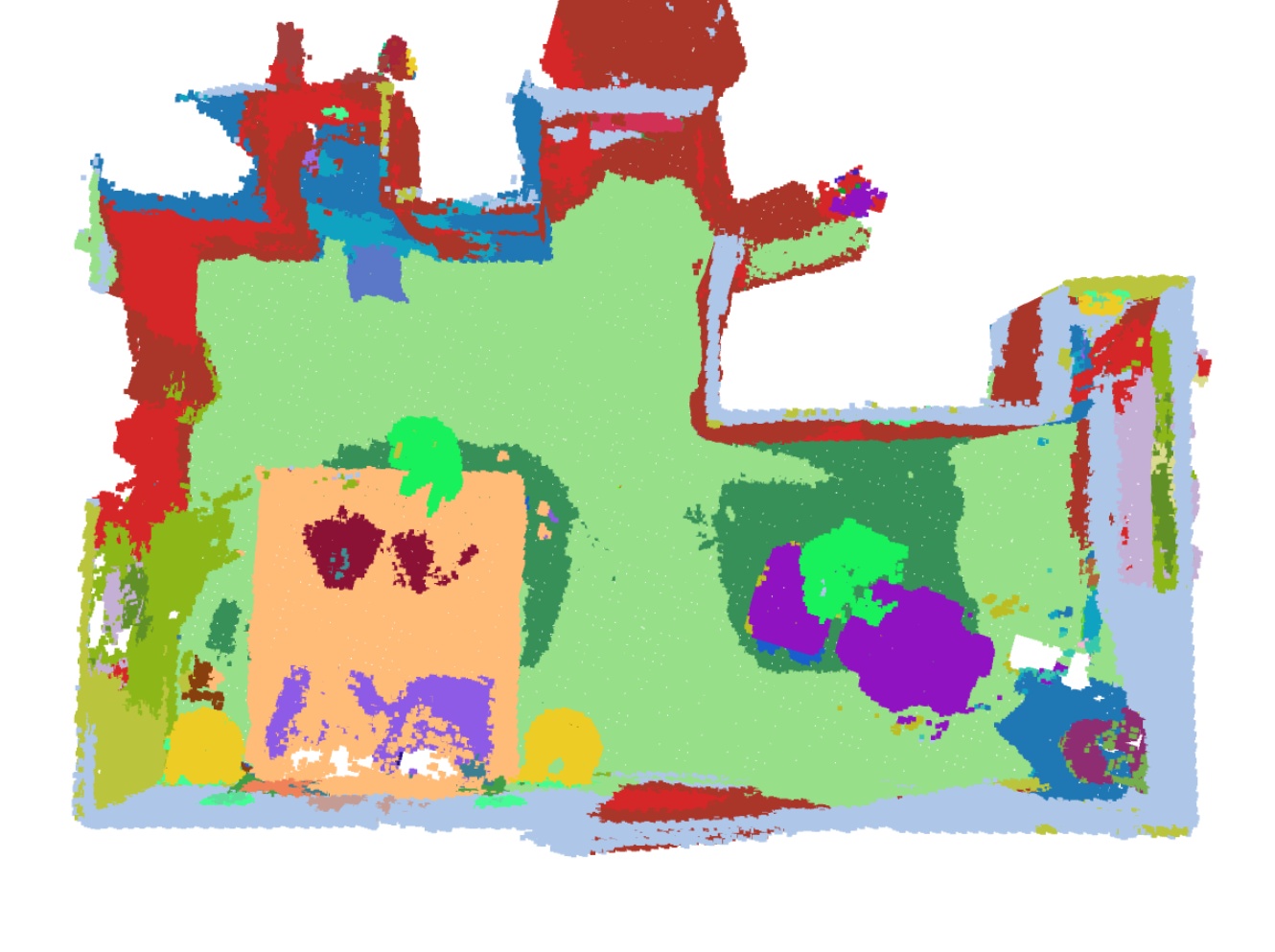} \\
    \includegraphics[width=0.24\linewidth]{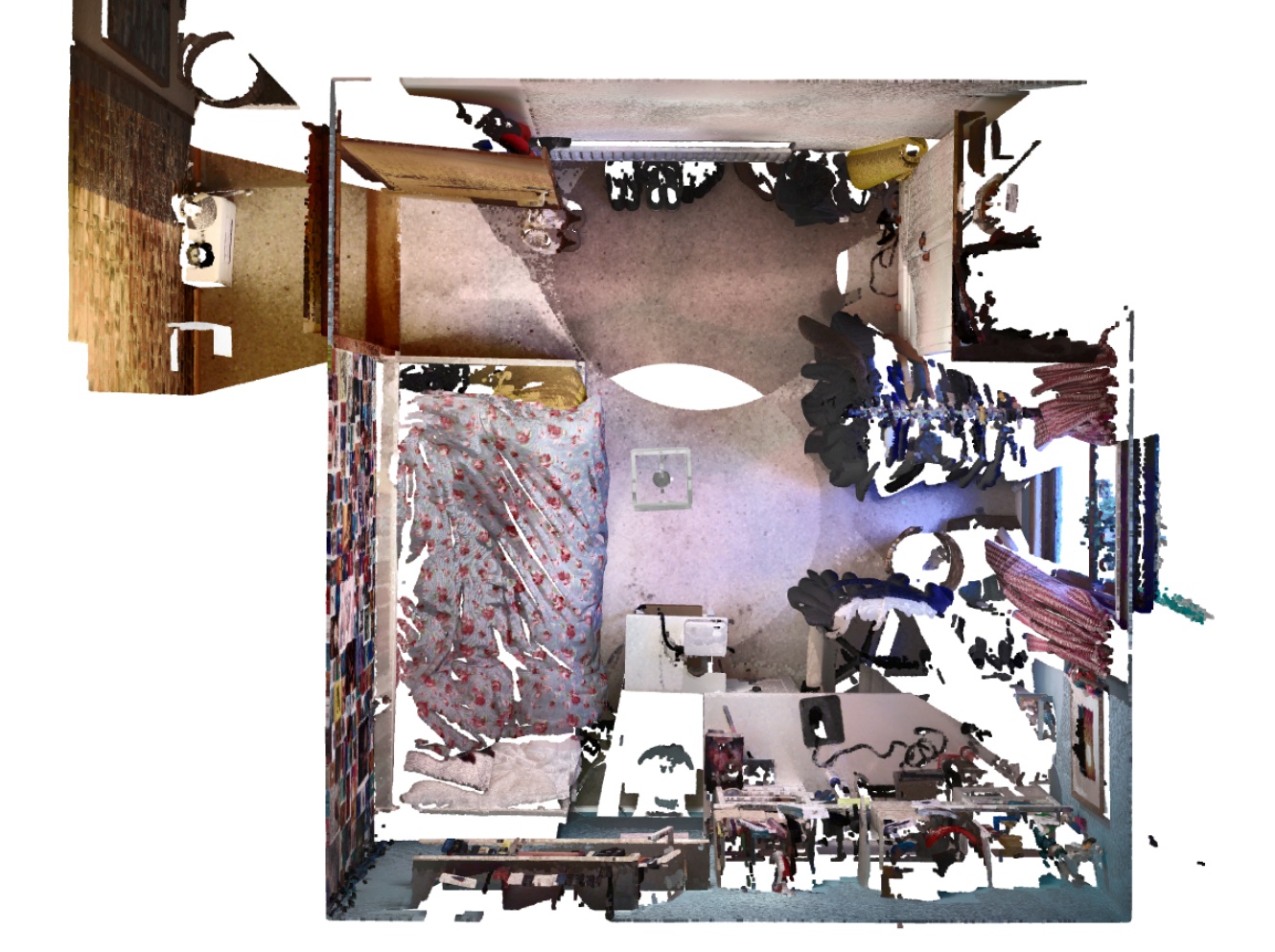}
    \includegraphics[width=0.24\linewidth]{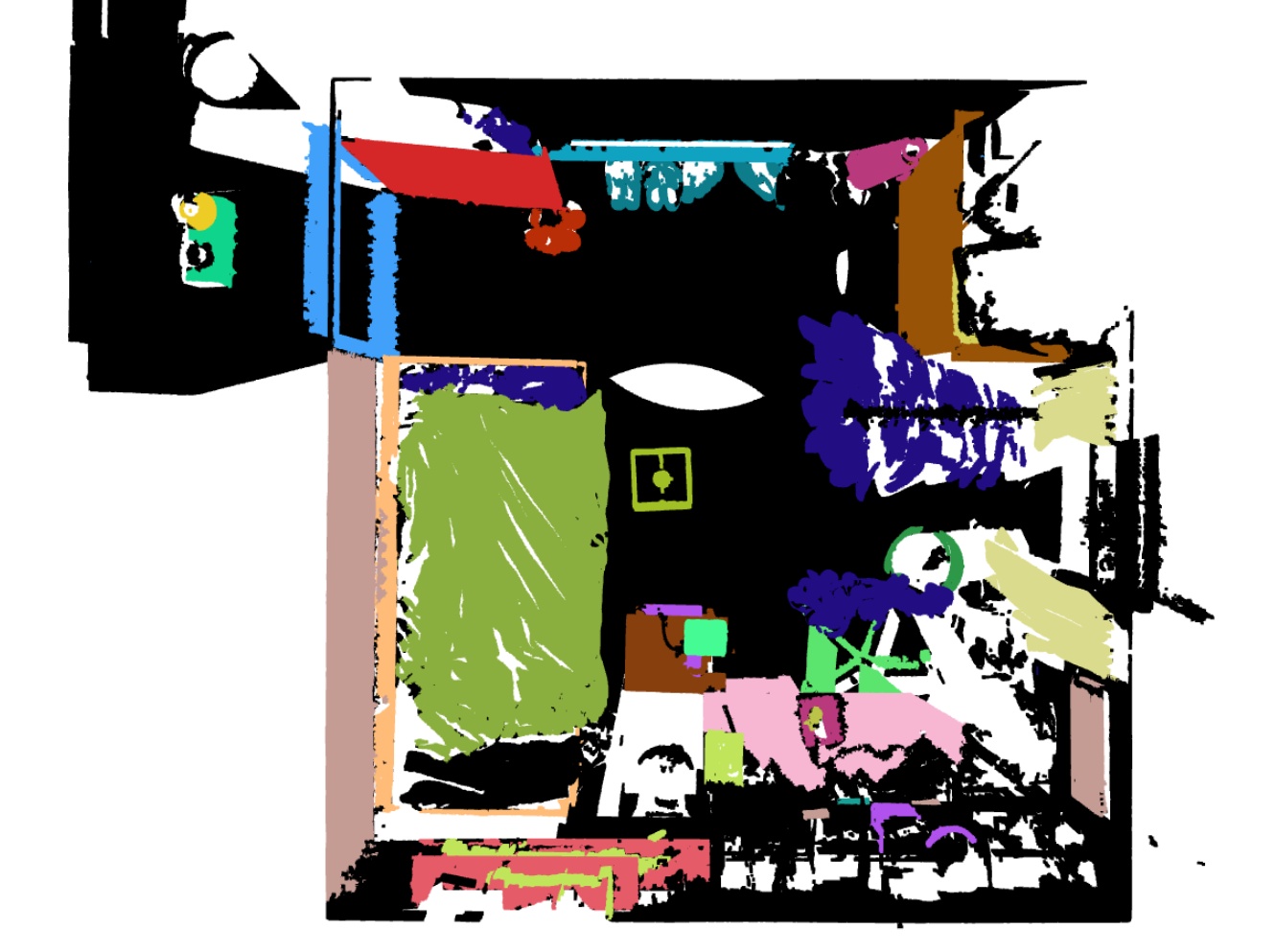}
    \includegraphics[width=0.24\linewidth]{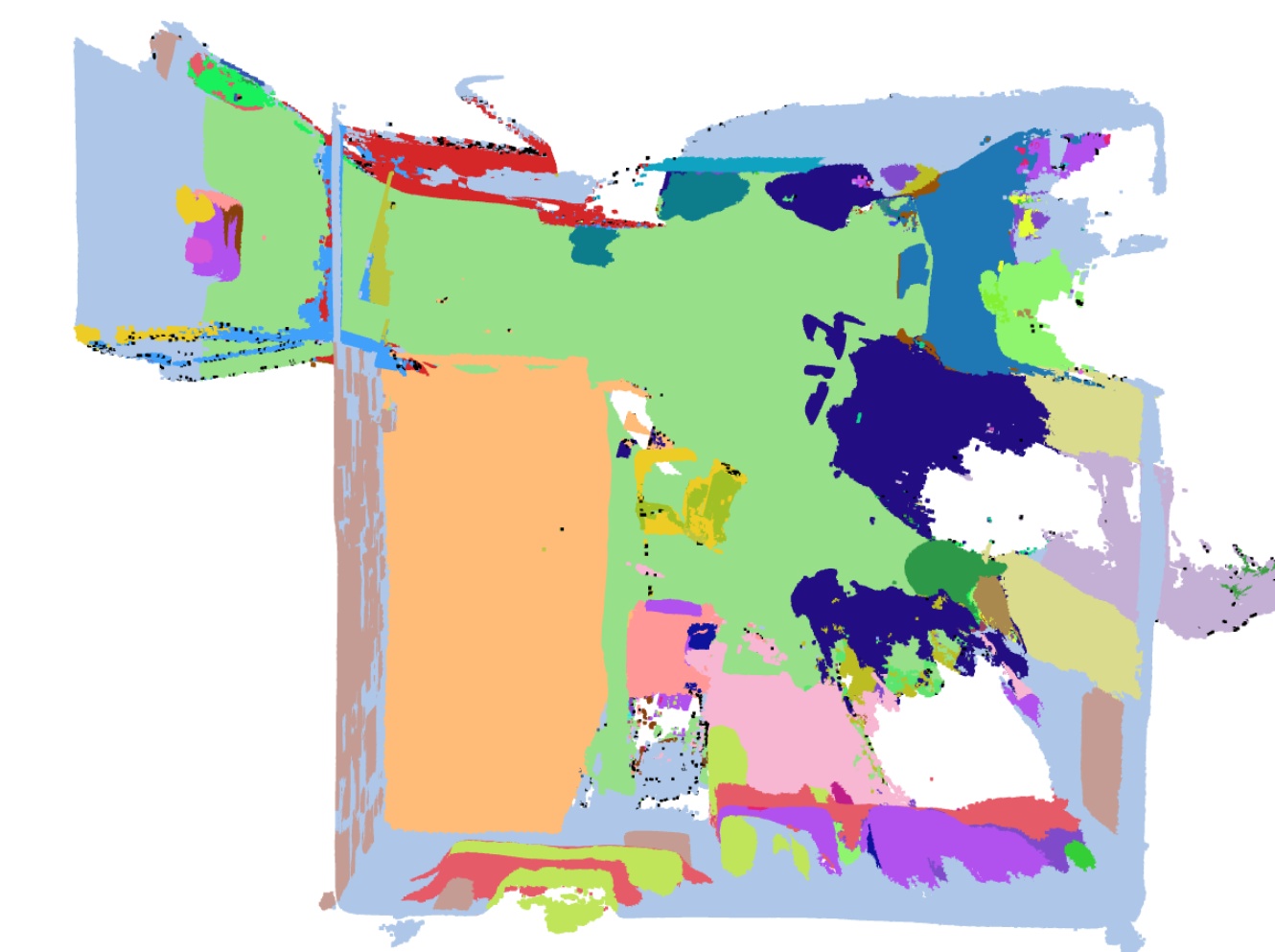}
    \includegraphics[width=0.24\linewidth]{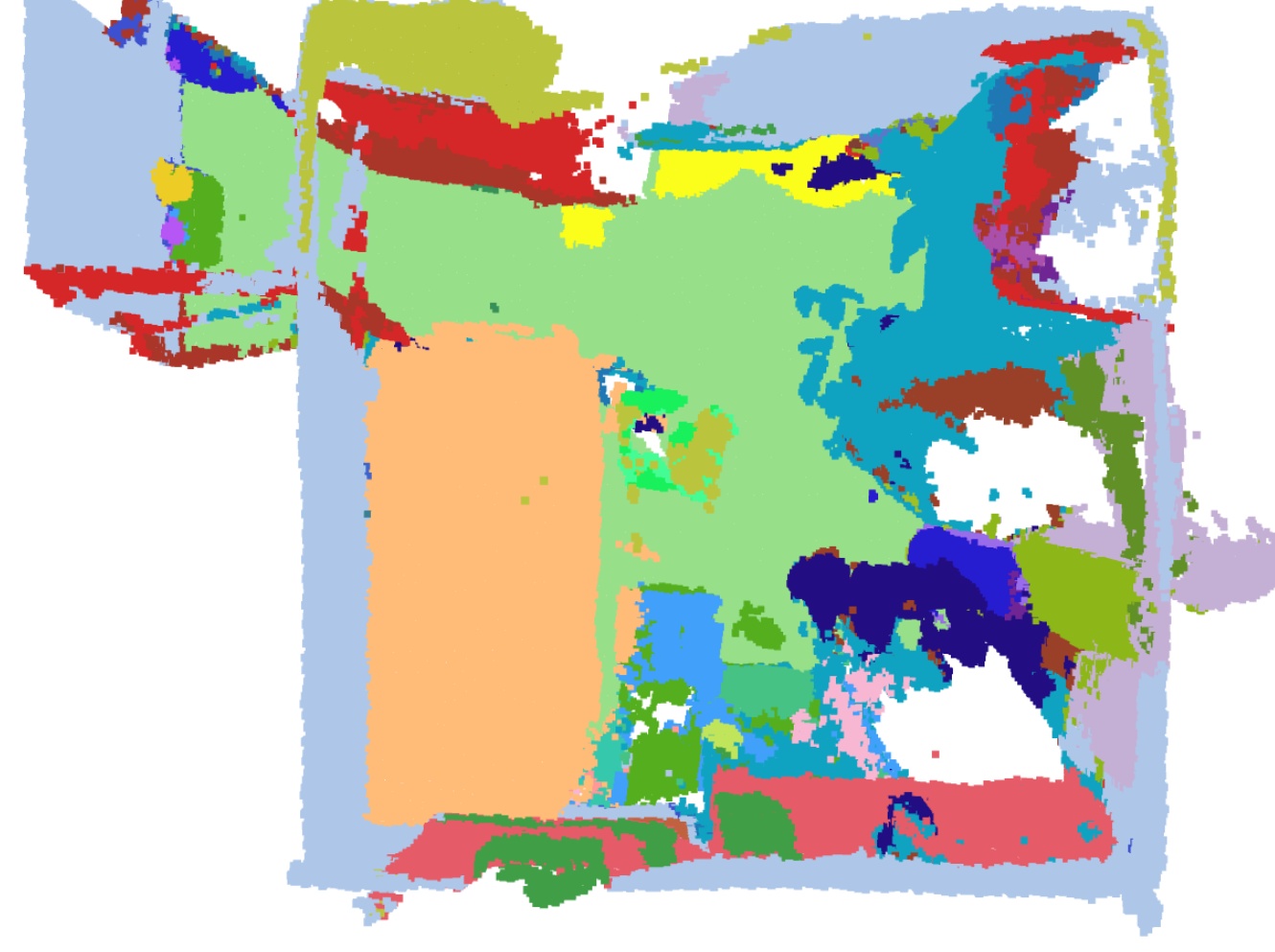} \\
    \includegraphics[width=0.24\linewidth]{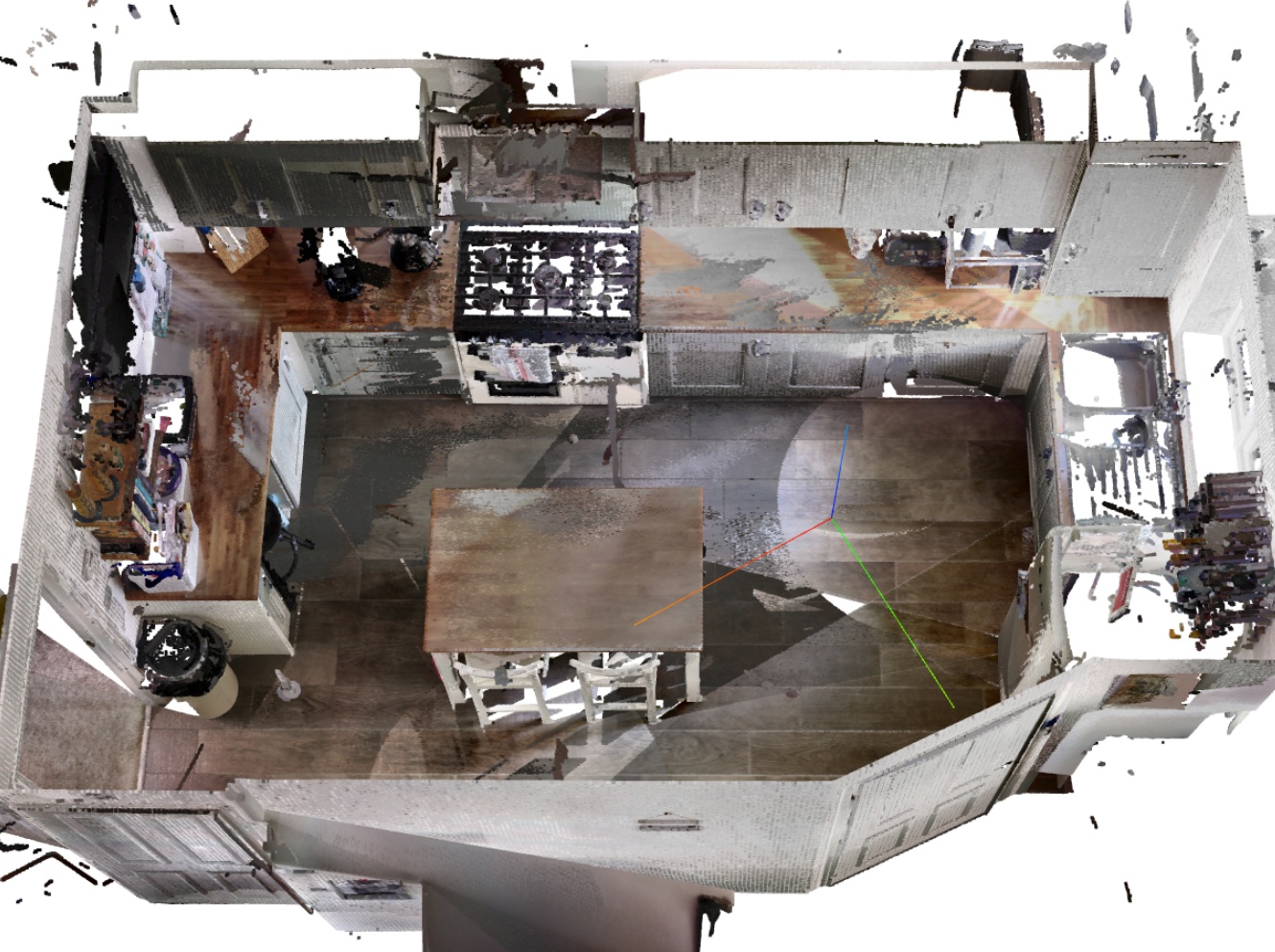}
    \includegraphics[width=0.24\linewidth]{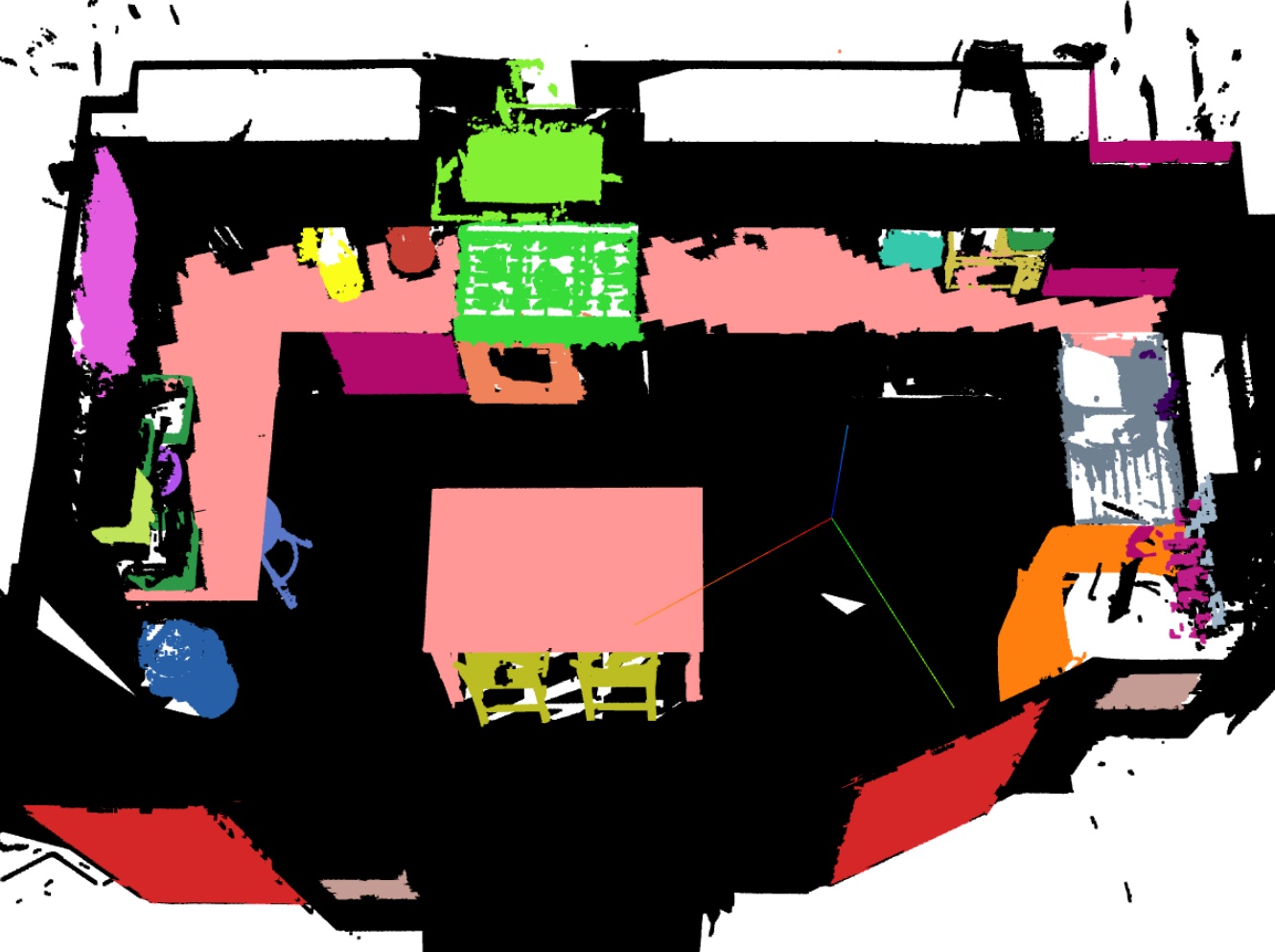}
    \includegraphics[width=0.24\linewidth]{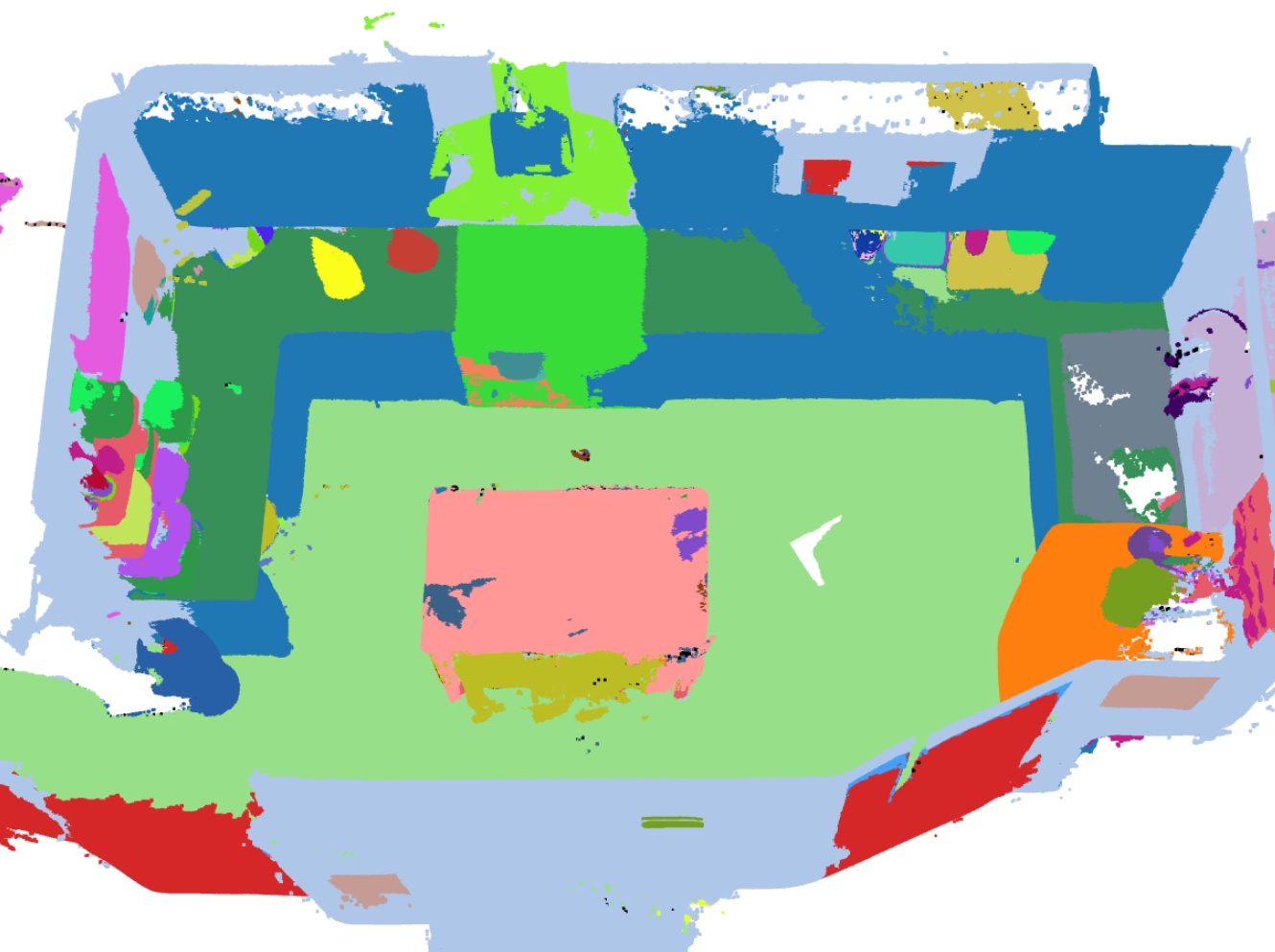}
    \includegraphics[width=0.24\linewidth]{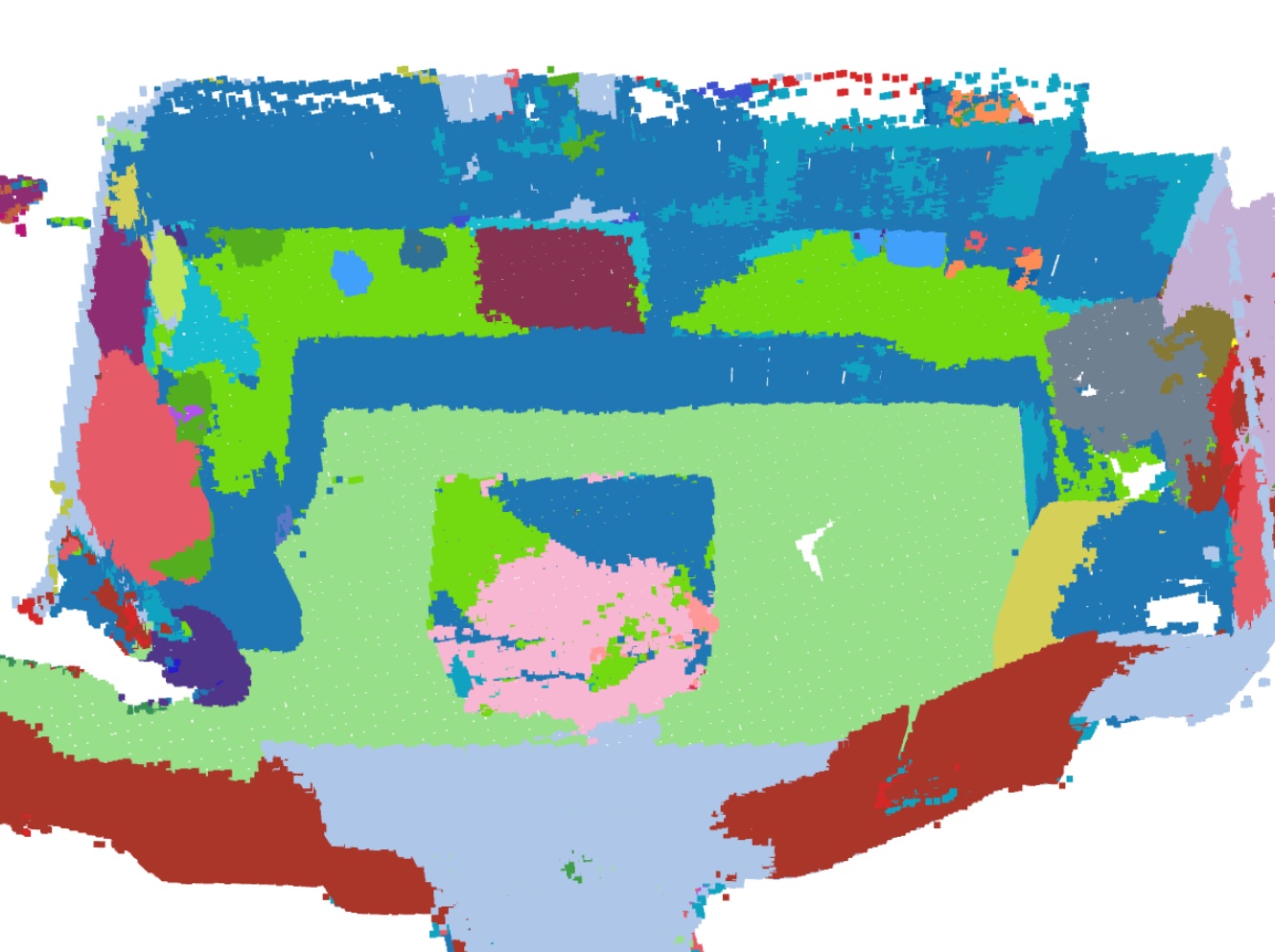} \\
    \caption{\label{reb:fig:qualitative_labels}\textbf{Visualization on ARKitScenes.}
        {From left to right}: 3D scene, ground truth annotation (black regions indicate unannotated areas), LabelMaker annotations, OpenScene predictions.}
\end{figure}

%% file: table_and_figure_tex/fig_5_mobile_capture_viz.tex
\begin{figure}[!t]
    \centering
    \includegraphics[width=0.49\linewidth]{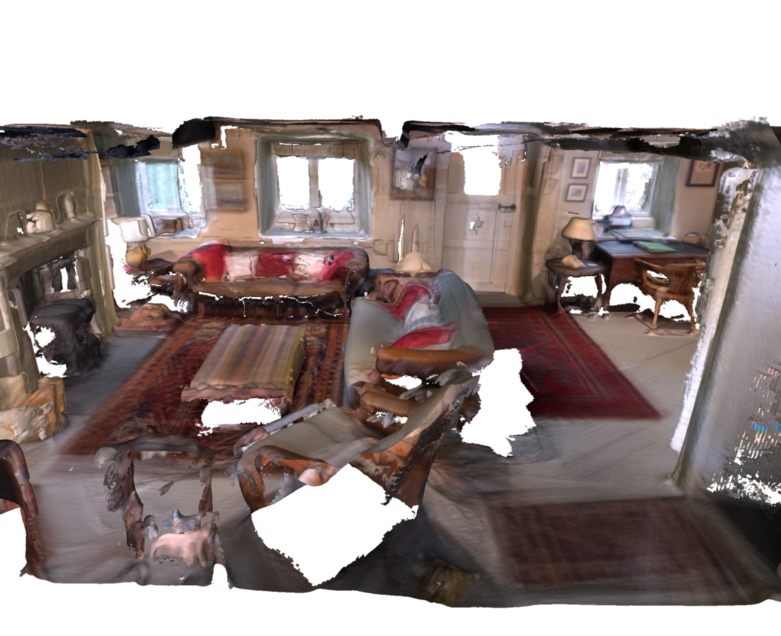}
    \includegraphics[width=0.49\linewidth]{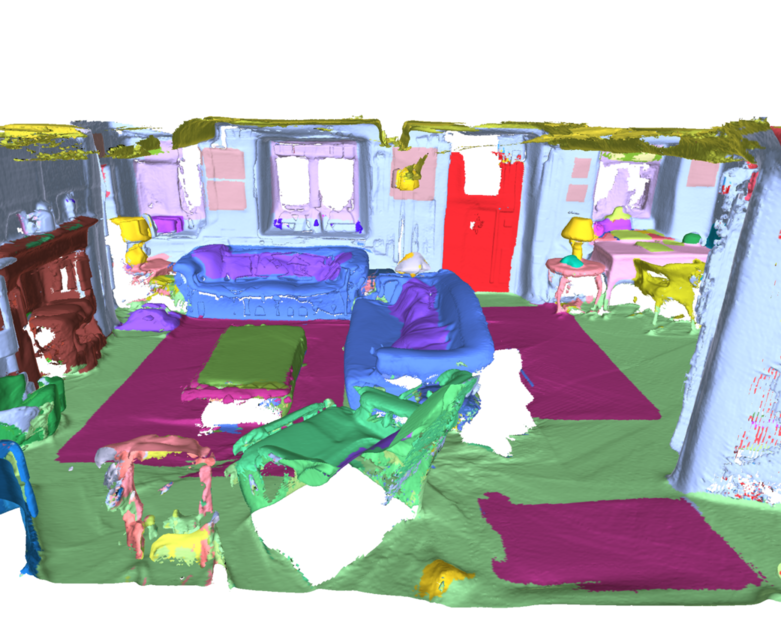} \\
    \includegraphics[width=0.49\linewidth]{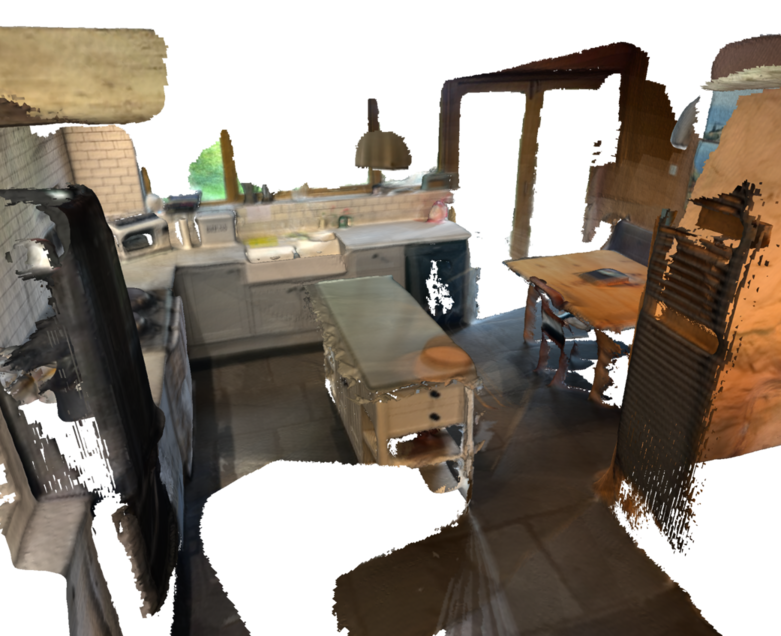}
    \includegraphics[width=0.49\linewidth]{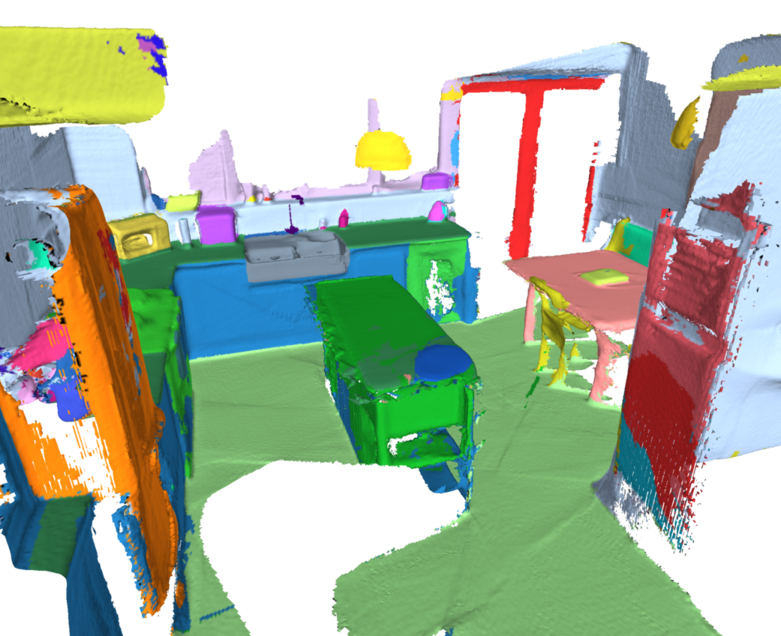}
    \caption{\textbf{Self-captured scenes and the semantic segmentation generated by LabelMakerv2}. This figure provides a qualitative impression of the segmentation quality obtained by LabelMakerV2. The color mapping is defined \href{https://github.com/cvg/LabelMaker/blob/main/labelmaker/mappings/label_mapping.csv}{here}.}
    \label{fig:viz}
\end{figure}

%% file: table_and_figure_tex/tab_7_matterport3d_zero_shot.tex
\begin{table*}[t!]
    \centering
    \footnotesize
    \resizebox{\linewidth}{!}{%
        \begin{tabular}{lccccccccc}
            \toprule
            label space                                                                  & \multicolumn{3}{c}{Top-40 NYU Classes} & \multicolumn{3}{c}{Top-80 NYU Classes} & \multicolumn{3}{c}{Top-160 NYU Classes}                                                                                                 \\\cmidrule(lr){2-4} \cmidrule(lr){5-7} \cmidrule(lr){8-10}
            method                                                                       & mIoU                                   & mAcc                                   & tAcc                                    & mIoU          & mAcc          & tAcc          & mIoU          & mAcc          & tAcc          \\\midrule
            MinkowskiNet~\cite{choy20194d} trained on Matterport3D (supervised)          & -                                      & 50.8                                   & -                                       & -             & 33.4          & -             & -             & 18.4          & -             \\
            MinkowskiNet trained on ScanNet200                                           & 33.1                                   & 43.5                                   & 74.5                                    & 19.9          & 28.4          & 72.3          & 11.7          & 17.5          & 71.5          \\
            MinkowskiNet trained on ARKit LabelMaker\textsuperscript{SN200} + ScanNet200 & 38.9                                   & 49.6                                   & 77.5                                    & 23.2          & 32.7          & 75.3          & 13.6          & 20.6          & 74.5          \\
            OpenScene\cite{openscene}                                                    & -                                      & 50.9                                   & -                                       & -             & 34.6          & -             & -             & 23.1          & -             \\
            vanilla PTv3 trained on ScanNet200                                           & 36.2                                   & 46.1                                   & 77.3                                    & 22.0          & 29.8          & 75.0          & 13.0          & 18.8          & 74.1          \\
            PTv3-PPT without ARKit LabelMaker (Our reproduction)                         & 41.4                                   & 51.2                                   & 79.9                                    & 27.0          & 36.2          & 77.7          & 16.3          & 24.4          & 76.8          \\
            PTv3-PPT trained with ARKit LabelMaker (Ours)                                & \textbf{43.8}                          & \textbf{53.6}                          & \textbf{80.6}                           & \textbf{29.1} & \textbf{38.2} & \textbf{78.5} & \textbf{17.3} & \textbf{26.0} & \textbf{77.6} \\
            \bottomrule
        \end{tabular}
    }
    \caption{\label{tab:matterport3d-ood}
        \textbf{Zero-shot evaluation on Matterport3D~\cite{Matterport3D} test set region segmentation}. We evaluate our trained PTv3 model on the ScanNet200 label space and map it to the top-40/80/160 NYU classes. Our models outperform fully-supervised MinkowskiNet and OpenScene.
    }
\end{table*}

%% file: table_and_figure_tex/tab_8_instance_seg.tex
\begin{table}[t!]
    \centering
    \resizebox{\linewidth}{!}{%
        \setlength{\tabcolsep}{2px}
        \begin{tabular}{lcccccc}
            \toprule
                                                                      & \multicolumn{3}{c}{\textbf{ScanNet}~\cite{dai2017scannet}} & \multicolumn{3}{c}{\textbf{ScanNet200}~\cite{scannet200}}                                                                 \\\cmidrule(lr){2-4} \cmidrule(lr){5-7}
            \textbf{Method}                                           & mAP$_{25}$                                                 & mAP$_{50}$                                                & mAP           & mAP$_{25}$    & mAP$_{50}$    & mAP           \\\midrule
            vanilla PTv3 PointGroup (from \cite{wu2024ptv3}           & 77.5                                                       & 61.7                                                      & 40.9          & 40.1          & 33.2          & 23.1          \\
            vanilla PTv3 PointGroup (reproduced)                      & 77.1                                                       & 62.9                                                      & 41.1          & 37.9          & 30.6          & 21.0          \\
            PTv3-PPT PointGroup pretrained \cite{wu2024ptv3}          & \textbf{78.9}                                              & \textbf{63.5}                                             & \textbf{42.1} & 40.8          & 34.1          & 24.0          \\
            PTv3-PPT PointGroup pretrained w/ ARKit LabelMaker (Ours) & 78.5                                                       & 62.6                                                      & 41.4          & \textbf{42.9} & \textbf{34.9} & \textbf{24.5} \\
            \bottomrule
        \end{tabular}
    }
    \caption{\label{tab:instance-seg}
        \textbf{Instance segmentation scores on ScanNet(200)~\cite{scannet200, dai2017scannet}}. Our dataset improves instance segmentation as a downstream task. Pre-training with our dataset increases performance on both ScanNet and ScanNet200, with a significant gain on ScanNet200.}
\end{table}

%% file: sec/5_conclusion.tex
\section{Conclusion}
\label{sec:conclusion}

In this paper, we introduced ARKit LabelMaker, the largest real-world 3D RGB-D dataset with 3D semantic annotations, generated automatically using an enhanced version of LabelMaker~\cite{labelmaker}, which we call LabelMakerV2. While these labels are inherently imperfect due to automation, we demonstrate their effectiveness in pre-training widely used 3D semantic segmentation models, significantly outperforming traditional, self-supervised, and augmentation-heavy training approaches. This aligns with trends in language and image generation, where scaling up training data has led to substantial performance gains. To support further data collection for training and evaluation, we also provide integration with an existing iOS app.

%% file: sec/X_suppl.tex
\clearpage
\setcounter{page}{1}
\setcounter{section}{0}
\setcounter{table}{0}
\setcounter{figure}{0}
\renewcommand{\thesection}{\Alph{section}}
\renewcommand{\thetable}{\Alph{section}\arabic{table}}
\renewcommand{\thefigure}{\Alph{section}\arabic{figure}}
\maketitlesupplementary

\section{Dataset Class Statistics}

\textbf{Dataset Statistics of ARKit LabelMaker. } In \Cref{fig:dataset_stats}, we present the point count for each class in the LabelMaker WordNet label space. Our dataset maintains a substantial data distribution even across tail classes.
\input{table_and_figure_tex/fig_s1_dataset_stats.tex}

\section{PTv3 Results on ScanNet++}

We also report the training and evaluation results of PTv3 on ScanNet++ in \Cref{tab:scannetpp-results}. Unfortunately, the numbers are not fully comparable, because we were so far unable to reporduce the validation results of PTv3. When the authors of \cite{wu2024ptv3} released PTv3's performance on ScanNet++, they expanded Structured3D’s training set from 6,519 to 18,348 samples, which we refer to as Structured3D v2. Due to limited computational resources, we could not train with this updated version of Structured3D yet. We will update ScanNet++ results once the new result is available. Our pre-training and joint-training (PPT) experiments show performance gains over vanilla PTv3, with PTv3-PPT achieving similar improvements to the original PTv3-PPT but with significantly less training data.

\input{table_and_figure_tex/tab_s1_scannetpp.tex}

\section{Tail Classes Performance of PTv3}

We give a detailed plot of the number of correctly predicted tail class points on ScanNet200 validation set in \Cref{fig:tail_class_bar_plot}.

\input{table_and_figure_tex/fig_s2_tail_class_bar_plot.tex}

\section{\label{sec:process}Detailed process of applying LabelMaker to ARKitScenes}

ARKitScenes is one of the largest indoor 3D scenes dataset. It consists of 5047 parsable scenes of various size.
We consider a scene parsable if the RGB-D trajecotry comes with associated pose data.
Processing these scenes with our improved LabelMaker pipeline requires deliberate engineering with the following goals: a) Bring the data in to the format required by LabelMaker~\cite{labelmaker} b) Robust processing to not waste compute on failures, c) Improved parallelization to speed up processing. d) Accurate resource estimation to prevent waste of compute resources and longer job waiting time. e) Fast failure identification and results inspection.

\textbf{Transforming the data}
LabelMaker~\cite{labelmaker} requires a specific data format to be able to reliably process all data. All trajectories require posed RGB-D data and a denoised 3D model that is used by Mask3D.
ARKitScenes comprises data of varying resolutions and sampling rates.
Depth data is captured at $256\times 192$ and 60 FPS, while the RGB frames are recorded at $640\times480$ and 30 FPS.
Therefore, synchronization is required to process the data with LabelMaker.
To this end, we match each RGB frame with the closest depth frame in time and we resize the depth frame to RGB frame's resolution.
Pose data, originally at 10 FPS, is interpolated using rotation splines to synchronize with each RGB frame.
To obtain a 3D mesh of each scene that can be processed by Mask3D, we reconstruct the 3D model by fusing the synchornized posed RGB-D data using TSDF fusion and then extract mesh with marching cube algorithm.
We empirically chose a voxel size of 8mm and a truncation distance of 4cm for fusion.

\textbf{Building a scalable pipeline}
LabelMaker~\cite{labelmaker} can be decomposed into individual modules such as the individual base models, the consensus computation, and the 3D lifting.
This modular nature allows to build a scalable pipeline using popular high-performance computing toolboxes.
As the different base models have different runtimes, we had to leverage dependency management system that can handle different dependencies of the pipeline steps.
This architecture allows us to effectively leverage large-scale computing and distribute the processing across many different nodes.

In more detail, we decompose the pipeline into several jobs (ordered by dependency) for each scene:
\begin{enumerate}[leftmargin=*]
    \item Preprocessing: Downloading the original scene data, transforming it into LabelMaker format, and run TSDF fusion to get the 3D mesh of the scene.
    \item Forwarding 2D images or 3D meshes to each base models: Grounded-SAM, Mask3D, OVSeg, CMX, InternImage. (all jobs depends on step 1.)
    \item Getting the consensus label from base models' labels. (depends on all elementary jobs in step 2.)
    \item Lifting the 2D consensus label into 3D. (depends on step 3.)
    \item Rendering the outputs of base models or consensus into videos for visualization. (depends on steps 2. or 3.)
    \item Post-processing, including removing temporary files and get statistics of each tasks. (depends on all steps above)
\end{enumerate}

\textbf{Optimizing compute resource scheduling.}
ARKitScenes contains scenes of a wide range of sizes, spanning from a minimum of 65 frames to a maximum of 13796, and different parts of the pipeline scale differently with increasing scene size.
To figure out the minimum resources requirements, we select 11 scenes of varied sizes uniformly distributed within the minimum and maximum range and record their resources usage.  While most jobs are not sensitive to scene size and can suffice with a fixed resource allocation, the base models exhibit greater sensitivity to scene size.
We interpolate resource needs with respect to scene size and summarize the empirical numbers into \Cref{tab:resource}.
Through this, we ensure that we request minimal-required resources, so that we have lowest job waiting time and less idle compute power.

\input{table_and_figure_tex/tab_s2_resource.tex}

\textbf{Assuring the quality of the individual processings.}
In order to assure high-quality labels produced by our improved pipeline, we have built tooling to efficiently check for failures of the processed scenes.
To this end, we store the logs and statistics of each job and built visualization tools for this data as well as the intermediate predictions.
This allows us to conveniently browse at scale through the predictions and identify individual failures.

\textbf{Failure handling and compute resource optimization.}
When doing large-scale processing on a high-performance compute cluster, a common issue is the failure of jobs.
This can happen for several reasons such as node crashing, unexpected memory usage, and many more.
Therefore, the processing pipeline has to be robust to these failures.
Additionally, compute should not be wasted when recovering from these failures.
Therefore, we designed a simple yet effective strategy for efficiently recovering from job failures.
Before every restart is triggered for a scene, we analyze both the logs and file system to identify the jobs that have not finished for this scene.
Once these jobs have been identified, we only resubmit these jobs.
This ensures that no compute resource is used in rerunning completed tasks.

\section{Log-Linear Performance Relation in Data Scaling}
\begin{figure}[th!]
    \centering
    \includegraphics[width=0.5\linewidth]{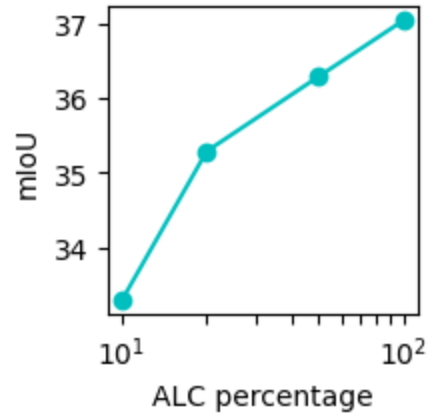}
    \caption{\label{reb:fig:scaling}\textbf{Relation of Validation mIoU against training data percentage of ARKit LabelMaker}. This figure shows the validation mIoU on ScanNet200 after fine-tuning with respect to the percentage of ARKit LabelMaker data used in pre-training. This figure shows a log-linear relationship.}
\end{figure}

\subsection{Implementation Details}

We optimize the code to deploy each individual piece of the pipeline of LabelMakerV2 as individual jobs to a GPU cluster, with SLURM as a dependency manager between the pipeline pieces. To optimize the overall execution time, it is therefore important to be able to estimate the processing time of each piece of the pipeline at the point of job submission.
ARKitScenes contains scenes of a wide range of sizes, spanning from a minimum of 65 frames to a maximum of 13796, and different parts of the pipeline scale differently with increasing scene size.
To figure out the minimum resources requirements, we select 11 scenes of varied sizes uniformly distributed within the minimum and maximum range and record their resources usage.  While most jobs are not sensitive to scene size and can suffice with a fixed resource allocation, the base models exhibit greater sensitivity to scene size.
We interpolate resource needs with respect to scene size and summarize the empirical numbers in the Appendix.
Through this, we ensure that we request minimal-required resources, so that we have lowest job waiting time and less idle compute power.

We use a CentOS 7 based SLURM cluster to process all the data, which is capable of handling task dependencies and parallel processing.
Before submitting jobs for a single scene, we first check the progress of each job and generate a SLURM script to submit only those unfinished jobs.
This ensures that no compute resource is used in rerunning completed tasks.

We employ test time augmentation by forwarding all models twice, with Mask3D using two different random seeds and other models being mirror flipped.
Following the practice of LabelMaker~\cite{labelmaker}, we assign equal weights to each model when calculating the consensus, although these weights are configurable in our pipeline code.
Since we are primarily interested in the 3D labels that can be used for pre-training 3D semantic segmentation models, SDFStudio training and rendering are omitted due to their lengthy processing times.
Further, we enhance the pipeline by storing both the most and second most voted predictions alongside their respective vote counts.
This information is useful when we want to investigate on the uncertainty across the base models.
We leave the exploitation of this information as potential future direction of research.

%% file: table_and_figure_tex/fig_s1_dataset_stats.tex
\begin{figure}[ht!]
    \centering
    \includegraphics[width=1.0\linewidth]{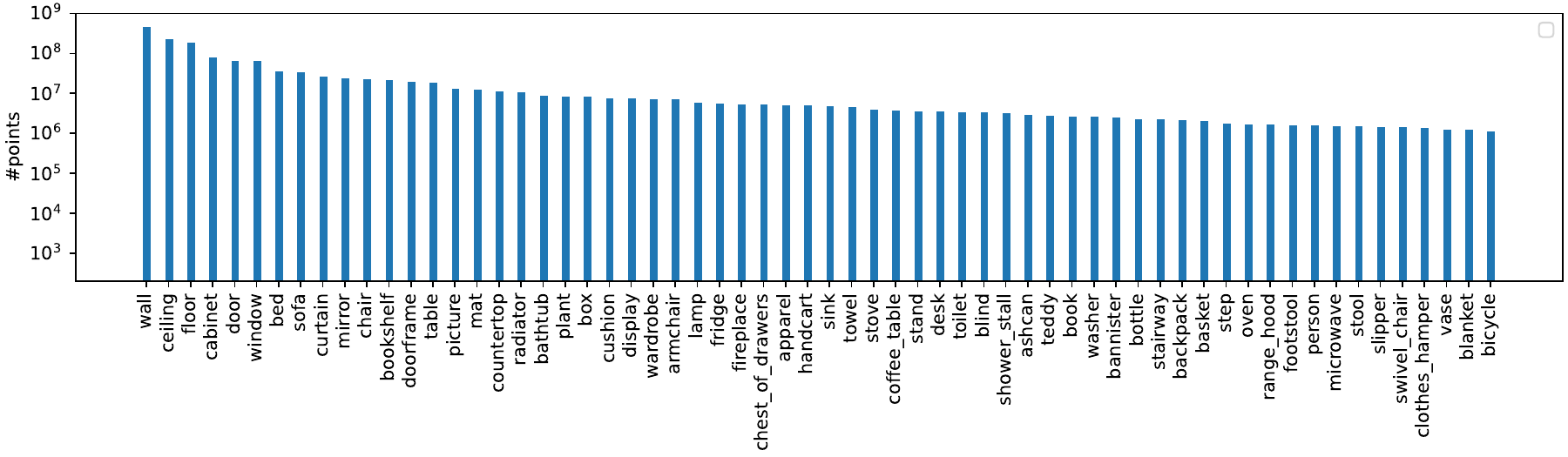}
    \includegraphics[width=1.0\linewidth]{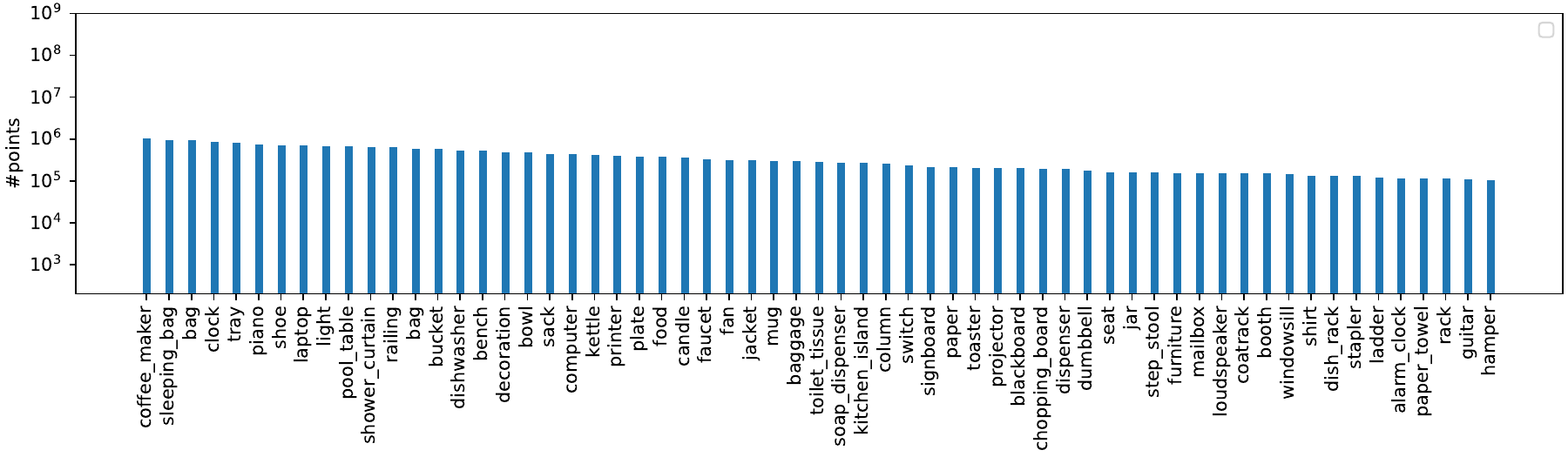}
    \includegraphics[width=1.0\linewidth]{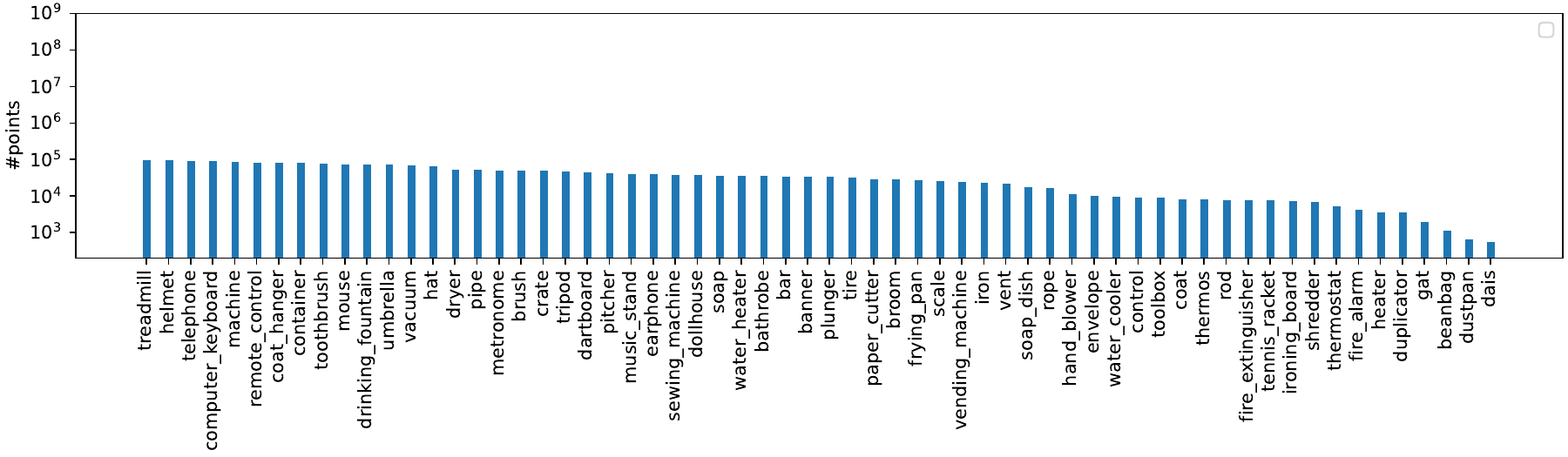}
    \caption{\textbf{Number of points for each ARKitLabelMaker class}.}
    \label{fig:dataset_stats}
\end{figure}

%% file: table_and_figure_tex/tab_s1_scannetpp.tex
\begin{table}[ht!]
    \centering
    \resizebox{\linewidth}{!}{%
        \begin{tabular}{lclccccc}
            \toprule
            \textbf{PTv3 Variant} & \textbf{Training Data}                                                            & \textbf{\#Data} & \textbf{val mIoU}    \\ \midrule
            vanilla               & ScanNet++                                                                         & 713             & 41.8                 \\
            fine-tune (Ours)      & ARKit LabelMaker\textsuperscript{SN200}  → ScanNet++                              & 4471 → 713      & 42.5                 \\
            PPT~\cite{wu2024ptv3} & {\scriptsize{ScanNet200 + ScanNet++ + Structure3Dv2}}                             & 45868           & \textbf{45.3}$^\dag$ \\
            PPT (Ours)            & {\scriptsize{ScanNet200 + ScanNet++  + ARKit LabelMaker}}                         & 11168           & 44.5                 \\
            PPT (Ours)            & {\footnotesize{ScanNet+ ScanNet200 + ScanNet++ + Structure3D + ARKit LabelMaker}} & 30386           & 44.6                 \\                                   \bottomrule
        \end{tabular}
    }
    \caption{\label{tab:scannetpp-results}\textbf{Semantic Segmentation Scores on ScanNet++~\cite{yeshwanthliu2023scannetpp}.}
        We evaluated the efficacy of our ARKit LabelMaker dataset on the ScanNet++ benchmark using both pre-training and joint training methods.
        $\dag$: this number comes from \citeauthor{wu2024ptv3}.
    }
\end{table}

%% file: table_and_figure_tex/fig_s2_tail_class_bar_plot.tex
\begin{figure}[ht!]
    \centering
    \includegraphics[width=1.0\linewidth]{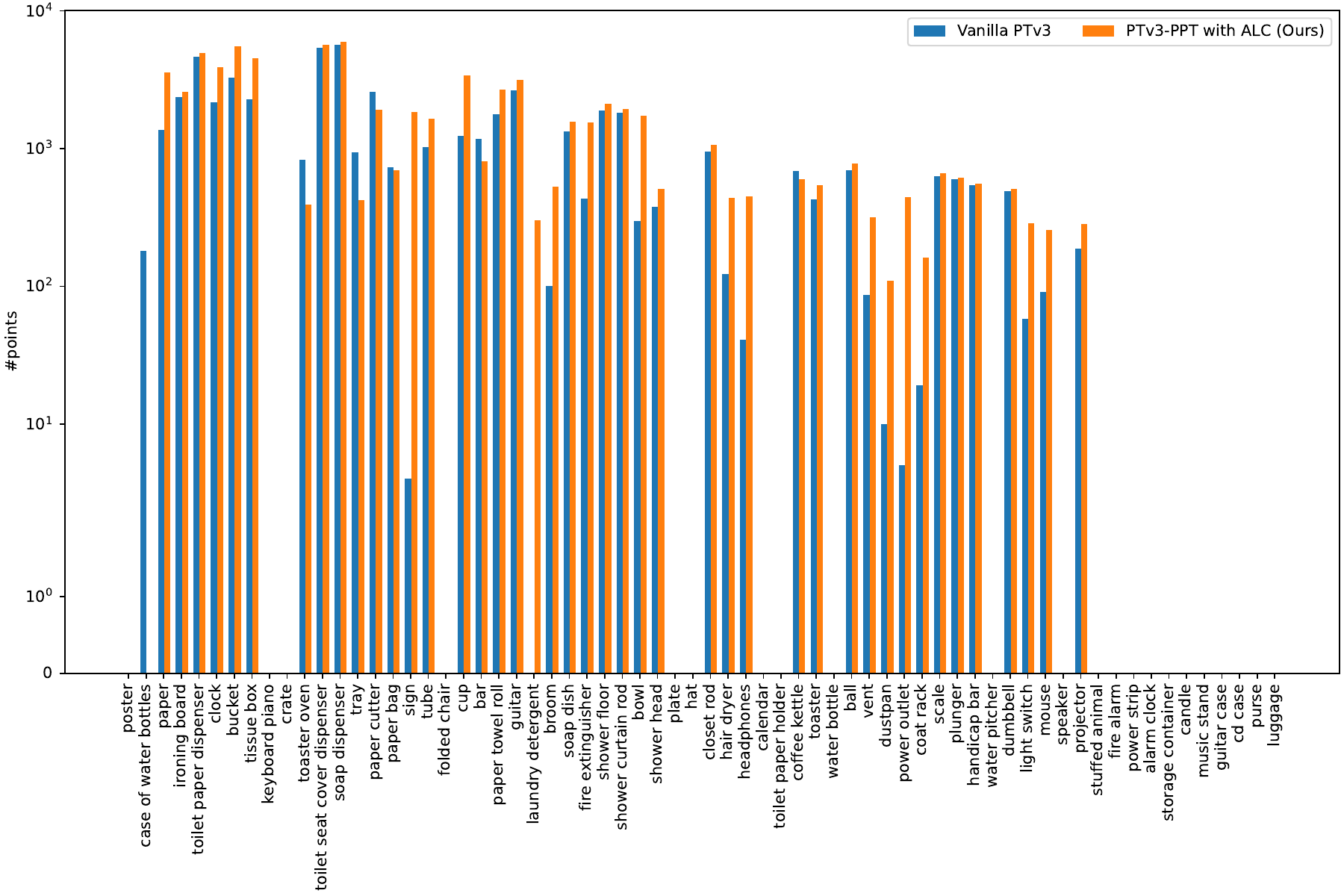}
    \caption{\textbf{Correctly predicted tail class points on ScanNet200 validation set.} We compare the number of correctly predicted points of tail class in ScanNet200 validation sets between PTv3 trained from scratch and the PTv3-PPT trained with our datasets. With our dataset, Point Transformer gains more ability to detect rase classes.}
    \label{fig:tail_class_bar_plot}
\end{figure}

%% file: table_and_figure_tex/tab_s2_resource.tex
\begin{table}[ht]
    \centering
    \resizebox{\linewidth}{!}{
        \begin{tabular}{l  c  c  c  c}
            \toprule
            Task                      & \#CPU & RAM & Time     & GPU             \\
            \midrule
            Download \& Prepossessing & 2     & 24G & 4h       & -               \\
            Video Rendering           & 8     & 32G & 30min    & -               \\
            Grounded-SAM              & 2     & 12G & 6h       & 3090  $\times1$ \\
            OVSeg                     & 2     & 8G  & 8h       & 3090 $\times1$  \\
            InternImage               & 2     & 10G & 8h       & 3090  $\times1$ \\
            Mask3D                    & 8     & 16G & 1h 30min & 3090 $\times1$  \\
            OmniData                  & 8     & 8G  & 2h       & 3090 $\times1$  \\
            HHA                       & 18    & 9G  & 2h       & -               \\
            CMX                       & 2     & 8G  & 3h       & 3090 $\times1$  \\ \midrule
            Consensus                 & 16    & 16G & 2h       & -               \\ \midrule
            Point Lifting             & 2     & 72G & 4h       & -               \\
            \bottomrule
        \end{tabular}
    }
    \caption{\label{tab:resource}\textbf{Requested resources for each task.}
        We report the average resources required by the individual steps of the LabelMakerv2 pipeline.
        The required cores, RAM, and GPU time varies across the different jobs.
        Through our job scheduling mechanism, we ensure that the required compute is optimially distributed across all jobs.
    }
\end{table}